\documentclass[10pt,twocolumn,letterpaper]{article}

\usepackage[pagenumbers]{cvpr} 

\usepackage[utf8]{inputenc} 
\usepackage[T1]{fontenc}    
\usepackage{url}            
\usepackage{booktabs}       
\usepackage{amsfonts}       
\usepackage{nicefrac}       
\usepackage{microtype}      
\usepackage[table,dvipsnames]{xcolor}
\usepackage{enumitem}
\usepackage{graphicx}
\usepackage{multirow}
\usepackage{pifont}
\usepackage{tcolorbox}
\usepackage{adjustbox}
\usepackage{amsmath,amssymb}
\usepackage{caption}
\usepackage{subcaption}
\usepackage{latexsym}
\usepackage{gensymb}
\usepackage{makecell}
\usepackage{marvosym}
\usepackage{tikz}
\usepackage{pgfplots}
\usepackage{xspace}
\usepackage{stfloats}
\usepackage{wrapfig}
\usepackage{cinzel}
\usepackage{tablefootnote}
\usepackage{footmisc}
\usepackage{footnote}
\usepackage[T1]{fontenc} 

\newcommand{\clipvit}{CLIP-ViT\xspace}

\newcommand{\ptvit}{pretrained-ViT\xspace}
\newcommand{\ptvits}{pretrained-ViTs\xspace}

\newcommand{\vit}{ViT\xspace}
\newcommand{\methodname}{{\scshape Vit-Lens}\xspace}
\newcommand{\Boldmethodname}{{\scshape \textbf{Vit-Lens}}\xspace}

\newcommand{\vitlensB}{{\scshape Vit-Lens}$_B$\xspace}
\newcommand{\vitlensL}{{\scshape Vit-Lens}$_L$\xspace}
\newcommand{\vitlensG}{{\scshape Vit-Lens}$_G$\xspace}

\newcommand{\projpage}{https://vit-lens.github.io}

\usepackage{mdframed}

\newcommand{\symbolHt}{1em}

\newcommand{\videoIcon}{%
  \begingroup\normalfont
  \includegraphics[height=\symbolHt]{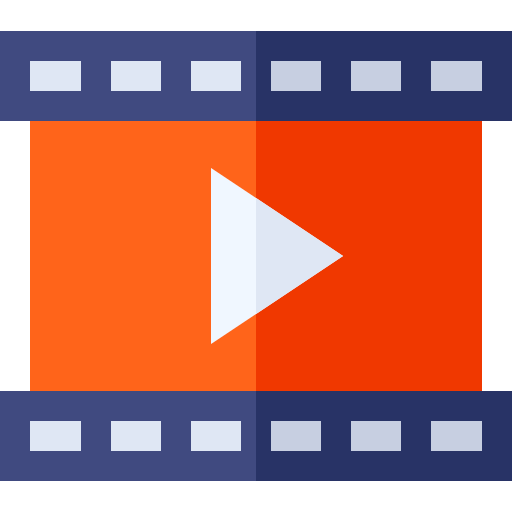}%
  \endgroup
}

\newcommand{\threeDIcon}{%
  \begingroup\normalfont
  \includegraphics[height=\symbolHt]{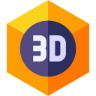}%
  \endgroup
}
\newcommand{\depthIcon}{%
  \begingroup\normalfont
  \includegraphics[height=\symbolHt]{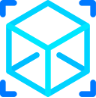}%
  \endgroup
}
\newcommand{\audioIcon}{%
  \begingroup\normalfont
  \includegraphics[height=\symbolHt]{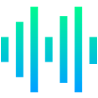}%
  \endgroup
}
\newcommand{\tactileIcon}{%
  \begingroup\normalfont
  \includegraphics[height=\symbolHt]{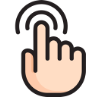}%
  \endgroup
}
\newcommand{\eegIcon}{%
  \begingroup\normalfont
  \includegraphics[height=\symbolHt]{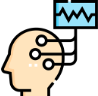}%
  \endgroup
}

\definecolor{pearThree}{HTML}{E74C3C}
\definecolor{pearDark}{HTML}{2980B9}
\definecolor{pearDarker}{HTML}{1D2DEC}
\definecolor{lavenderblue}{rgb}{0.8, 0.8, 1.0}
\definecolor{lavenderweb}{rgb}{0.9, 0.9, 0.98}
\definecolor{mypurple}{RGB}{200,192,248}
\definecolor{mypurpledeep}{RGB}{142,126,240}
\definecolor{mygreen}{RGB}{117,170,156}
\definecolor{myyellow}{RGB}{255,192,0}
\definecolor{myblue}{RGB}{57,143,255}
\definecolor{mygrey}{RGB}{231,230,230}
\definecolor{codey}{RGB}{220,220,170}
\definecolor{coder}{RGB}{206,145,120}
\definecolor{codeb}{RGB}{156,220,254}
\definecolor{codenum}{RGB}{204,204,204}
\definecolor{ADark}{rgb}{0,0.3,0.8}
\definecolor{BDark}{rgb}{.5,.0,.5}
\definecolor{CDark}{rgb}{0,.5,0}
\definecolor{DDark}{rgb}{0.11764705882352941, 0.5647058823529412, 1.0}
\definecolor{EDark}{rgb}{0.8823529411,0.63725490196,0.0156862745}
\definecolor{FDark}{rgb}{0.6235294117647059, 0.27058823529411763, 0.4627450980392157}

\definecolor{Gray}{gray}{0.92}
\definecolor{DarkGray}{gray}{0.5}

\newcommand{\cmark}{\textcolor{mygreen}{\ding{51}}} 
\newcommand{\xmark}{\textcolor{BrickRed}{\ding{55}}} 

\newcommand{\dsA}{\textcolor{myyellow}{$\blacktriangleright$}}
\newcommand{\dsB}{\textcolor{mygreen}{$\blacktriangleright$}}
\newcommand{\dsC}{\textcolor{mypurpledeep}{$\blacktriangleright$}}

\newcommand{\head}[1]{{\noindent\textbf{#1}}}
\newcommand{\ve}[1]{\mathbf{#1}}

\definecolor{cvprblue}{rgb}{0.21,0.49,0.74}
\usepackage[pagebackref,breaklinks,colorlinks,citecolor=cvprblue]{hyperref}

\def\paperID{****} 
\def\confName{CVPR}
\def\confYear{2024}

\title{\includegraphics[scale=0.032, bb=-100 8 300 34]{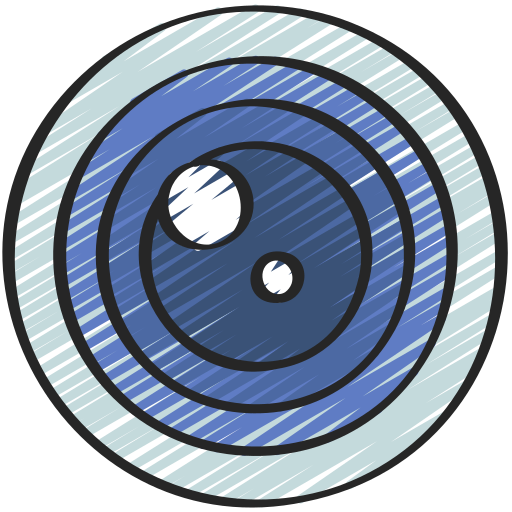}
\methodname: Towards Omni-modal Representations}

\author {
    Weixian Lei\textsuperscript{\rm 1,2}\quad
    Yixiao Ge\textsuperscript{\rm 2,3}$^{\dag}$\quad
    Kun Yi\textsuperscript{\rm 2}\quad
    Jianfeng Zhang\textsuperscript{\rm 1}\quad
    Difei Gao\textsuperscript{\rm 1}\\
    Dylan Sun\textsuperscript{\rm 2}\quad
    Yuying Ge\textsuperscript{\rm 3}\quad
    Ying Shan\textsuperscript{\rm 2,3}\quad
    Mike Zheng Shou\textsuperscript{\rm 1}$^{\dag}$\\
    \small $^{\dag}$Corresponding authors \\
    \textsuperscript{\rm 1}Show Lab, National University of Singapore\quad
    \textsuperscript{\rm 2}ARC Lab, Tencent PCG\quad
    \textsuperscript{\rm 3}Tencent AI Lab
}
\begin{document}

\twocolumn[{%
\renewcommand\twocolumn[1][]{#1}%
\maketitle
\begin{center}
    \centering
    \captionsetup{type=figure}
    \includegraphics[width=1.\linewidth]{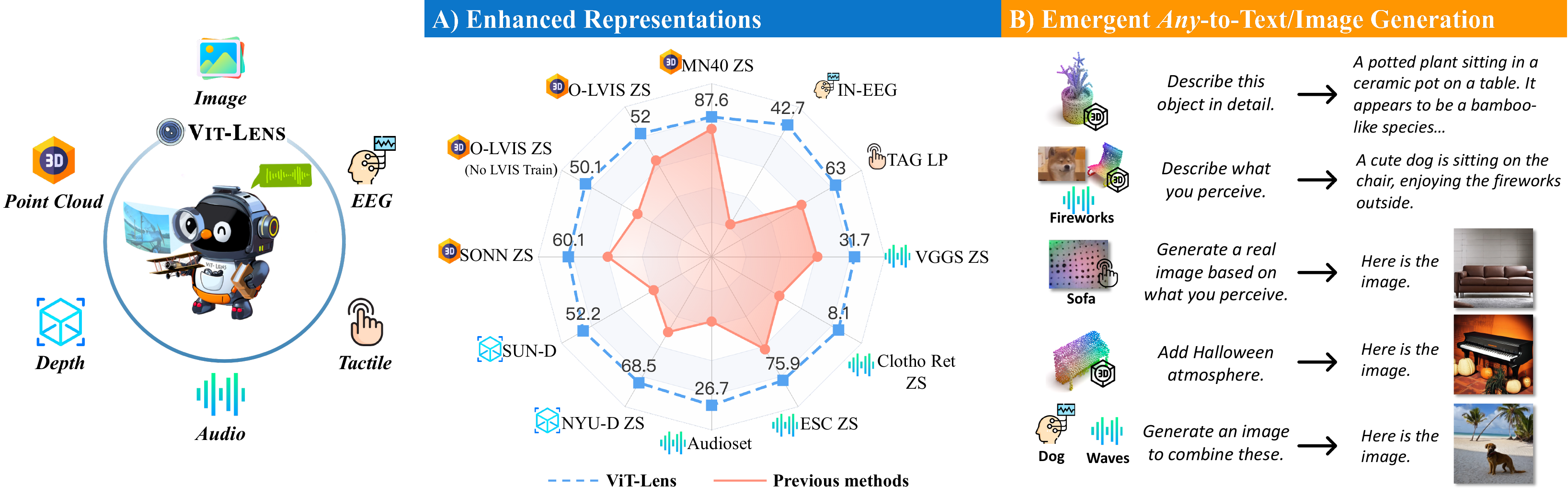
    }
    \captionof{figure}{
    \textbf{\methodname for omni-modal representation learning. }
    \textbf{A)} \methodname consistently enhances the performance of understanding tasks, such as classification, zero-shot classification (ZS) and linear probing (LP), across 3D point cloud(\cite{liu2023openshape}), depth(\cite{girdhar2023imagebind}), audio(\cite{girdhar2023imagebind}), tactile(\cite{yang2022touch_and_go}), and EEG(\cite{bai2023dreamdiffusion}) modalities. The citations represent the compared previous methods. Further details in~\cref{sec:experiments}.
    \textbf{B)} By plugging \methodname into multimodal foundation models, it enables emergent applications ``out-of-the-box'', including Any-modality Captioning/QA, Any-modality-to-Image Generation and text-guided Any-modality-to-Image editing, to name a few. 
}
\label{fig:teaser}
\end{center}%
}]

\makeatletter{\renewcommand*{\@makefnmark}{}
\footnotetext{$^*$This work extends~\cite{lei2023vitlens} with added modalities and applications.}\makeatother}

\begin{abstract}
Aiming to advance AI agents, large foundation models significantly improve reasoning and instruction execution, yet the current focus on vision and language neglects the potential of perceiving diverse modalities in open-world environments. 
However, the success of data-driven vision and language models is costly or even infeasible to be reproduced for rare modalities.
In this paper, we present \Boldmethodname that facilitates efficient omni-modal representation learning by perceiving novel modalities with a \ptvit and aligning them to a pre-defined space.
Specifically, the modality-specific lens is tuned to project any-modal signals to an intermediate embedding space, which are then processed by a strong ViT with pre-trained visual knowledge.
The encoded representations are optimized toward aligning with the modal-independent space, pre-defined by off-the-shelf foundation models. 
\methodname provides a unified solution for representation learning of increasing modalities with two appealing advantages:
(i) Unlocking the great potential of \ptvits to novel modalities effectively with efficient parameters and data regime;
(ii) Enabling emergent downstream capabilities through modality alignment and shared \vit parameters.
We tailor \methodname to learn representations for 3D point cloud, depth, audio, tactile and EEG, and set new state-of-the-art results across various understanding tasks, such as zero-shot classification.
By seamlessly integrating \methodname into Multimodal Foundation Models, we enable Any-modality to Text and Image Generation in a zero-shot manner. 
Code and models are available at \url{https://github.com/TencentARC/ViT-Lens}.
\end{abstract}    
\section{Introduction}
\label{sec:intro}
Humans interact with the world through various sensory systems like vision, audition, touch, smell, and taste. To advance versatile AI agents, deep learning models need to replicate these human-like multi-sensory abilities and tackle varied user-specified tasks.
For instance, visually interpreting road signs to ensure our safe driving, listening to sirens to respond to emergency vehicles, and tactually assessing clothing fabric quality to offer shopping guidance.
Among these applications, omni-modal representation learning has become a focal point, which enables comprehensive perception in open-world environments.

On the way to pursuing omni-modal AI agents, the research community has utilized large-scale web data to make substantial strides in language~\cite{bert,roberta,gpt1,gpt2,gpt3,openai2022chatgpt,llama} and vision~\cite{dosovitskiy2020image,he2022masked,bao2021beit,peng2022beitv2,fang2023eva,fang2023eva02,openai_clip,jia2021align,cherti2022openclip,yu2022coca}.
Consequently, Multimodal Foundation Models (MFMs)~\cite{alayrac2022flamingo,zhu2022minigpt4,liu2023llava,chen2022pali,driess2023palm,dai2023instructblip,ge2023seed_tokenizer,ge2023seed_llama} that integrate vision representations with Large Language Models (LLMs) have made great progress in both vision-language comprehension and generation.

However, extending the success of unleashing LLMs to comprehend and interact with a broader array of modalities remains challenging.  
Despite recent initiatives~\cite{girdhar2023imagebind,wang2023onepeace} in pursuing omni-modal intelligence, their capabilities on certain modalities are often constrained by limited data used in the training phase.
In contrast to image, video, and text data, which are abundant on the internet, acquiring large-scale datasets for less common modalities can be non-trivial. This scarcity of data leads to sub-optimal models with poor generalization, particularly when encountering novel categories, thereby limiting their broader real-world applications.

In this work, we present a novel perspective. 
Given the exceptional generalization and transfer learning capabilities of the \ptvits~\cite{openai_clip,fang2023eva,fang2023eva02,caron2021dino,oquab2023dinov2}, there is promise in adapting their inherent knowledge to comprehend novel modalities.
This eliminates the necessity of collecting large-scale data to train models from scratch for each modality, which demands substantial time and resources. 
Recognizing the rich knowledge encoded in a \ptvit, we conjecture that a \ptvit is able to function as a multi-modal processor -- it possesses the capacity to sense and comprehend a spectrum of modalities as it interprets images.

From this standpoint, 
we introduce \Boldmethodname, which encodes the out-of-image modalities through a set of \ptvit parameters, with the goal of maximizing the utilization of pretrained model weights and the knowledge they encapsulate.
Specifically, \methodname employs a modality-specific Lens along with a lightweight modality embedding module to transform input data into an intermediate space. Subsequently, a frozen \ptvit is applied for further encoding. 
This approach enables the encoding of diverse modalities, aligning their features with the established features of anchor data, which can range from images, text to others, from off-the-shelf foundation models.

Our proposed method offers several advantages in advancing omni-modal representation learning:
{\bf (1) Parameters and data efficient approach.} Our method adopts a shared set of \ptvit parameters across various modalities, enabling an efficient utilization of model parameters.  Moreover, it efficiently enhances representations for less common modalities by leveraging the advanced \vit model, reducing the demand for extensive data collection.
{\bf (2) Emergent capability.} By training \methodname with the \vit used in an off-the-shelf MFM, we can seamlessly obtain an Any-Modality MFM via \methodname integration.
The integrated model extends the original MFM's capabilities to various modalities, without any specific instruction tuning.
For instance, without direct training for tactile data, the model broadens its image generation capability to include tactile-to-image generation. As a result, it can generate an image of a sofa upon receiving tactile signals indicating ``leather''.

We conducted comprehensive experiments across multiple modalities, extending beyond images and videos to encompass 3D point cloud, depth, audio, tactile, and EEG. This experiments were evaluated across 11 benchmarks.
As is shown in~\cref{fig:teaser}\textcolor{red}{A}, \methodname demonstrates state-of-the-art performance in 3D zero-shot classification. Particularly, when LVIS classes are excluded during training, \methodname achieves an impressive zero-shot classification accuracy of 50.1\% on Objaverse-LVIS~\cite{deitke2023objaverse}, surpassing the prior SOTA by 11.0\%.
It consistently outperforms ImageBind~\cite{girdhar2023imagebind} on depth and audio benchmarks, and surpasses previous works on tactile~\cite{yang2022touch_and_go} and EEG~\cite{bai2023dreamdiffusion} related tasks.

Beyond understanding tasks, we plug \methodname into two recent MFMs, InstructBlip~\cite{dai2023instructblip} and SEED~\cite{ge2023seed_tokenizer, ge2023seed_llama}. 
As illustrated in~\cref{fig:teaser}\textcolor{red}{B}, this empowers the MFMs to comprehend any modality in a zero-shot manner, making Any-Captioning, Any-QA, Any-to-Image Generation and text guided Any-to-Image editing right out of the box, all without the need for specific instruction tuning.

\section{Related Work}
\label{sec:related_work}
\par\head{Vision Language Pretraining: Advancements and Impacts.}
Recent advancements in vision-language pretraining, including models such as CLIP~\cite{openai_clip}, ALIGN~\cite{jia2021align}, CoCa~\cite{yu2022coca}, Flamingo~\cite{alayrac2022flamingo}, and LiT~\cite{zhai2022lit}, have leveraged image-text pairs to achieve remarkable zero-shot performance on a wide range of vision and language tasks. Meanwhile, pretrained CLIP models have served as influential teachers and their joint embedding space has demonstrated efficacy in diverse zero-shot tasks such as segmentation~\cite{li2022languagedriven}, detection~\cite{gu2021open,zhou2022detecting}, 3D shape understanding~\cite{xue2023ulip,xue2023ulip2,liu2023openshape,zhang2022pointclip,zhu2022pointclipv2}, 3D open-vocabulary segmentation~\cite{Peng2023OpenScene}, mesh animation~\cite{youwang2022clipactor}, audio understanding~\cite{guzhov2022audioclip} and more~\cite{zhang2023magicavatar}. \methodname extends these models' capacities to diverse modalities by integrating \ptvit, enhancing its omni-modal understanding ability and enabling superior performance across various tasks and modalities.

\par\head{Multimodal Learning.}
Previous studies explored joint training across multiple modalities in both supervised~\cite{girdhar2022omnivore,likhosherstov2021polyvit, gao2020multi} and self-supervised settings~\cite{girdhar2023omnimae,arandjelovic2017look,tian2020contrastive,morgado2021audio, lin2022egocentric}.
Several approaches aim at aligning various modalities to CLIP for multimodal zero-shot learning. AudioCLIP~\cite{guzhov2022audioclip} adds audio to CLIP for zero-shot audio classification, while ImageBind~\cite{girdhar2023imagebind} aligns six modalities to CLIP using paired image data.
Besides, ONE-PEACE~\cite{wang2023onepeace} introduces a unified encoder that is pretrained from scratch to align vision, language, and audio. 
Zhang \etal~\cite{zhang2023metatransformer} pretrain a transformer with LAION-2B, following CLIP's methodology, for downstream supervised tasks across modalities.
In contrast, \methodname stands out by leveraging a \ptvit to understand and unite diverse modalities without manual annotations. Its seamlessly integration with Multimodal Foundation Models (MFM) allows easy plug-and-play in emergent applications.

\par \head{Multimodal Foundation Models.}
Recent advancements in Large Language Models (LLMs)~\cite{openai2022chatgpt,llama} have demonstrated remarkable language understanding and reasoning abilities. Afterwards, substantial efforts~\cite{alayrac2022flamingo,zhu2022minigpt4,liu2023llava,liu2023llava1.5,li2023videochat} have been directed towards enabling LLMs to perceive and interact with the visual world with the help of visual representation models. Similar paradigms enable LLMs to understand more modalities by algning the well-trained encoders of various modalities to the textual space of LLMs~\cite{su2023pandagpt,han2023imagebindllm,guo2023pointbind}.
Beyond understanding tasks, recent works~\cite{Emu,ge2023seed_tokenizer,ge2023seed_llama} empower LLMs with the ability to generate images, and NextGPT~\cite{wu2023nextgpt} extends the generative capabilities to encompass audio and video.
Most of these Multimodal Foundation Models (MFMs) require specific instruction-following data within particular domains for training. In this study, we demonstrate that \methodname can seamlessly integrate with an MFM without additional training, extending its capabilities to various modalities.
\section{Method}
\begin{figure}[t]
  \centering
  \includegraphics[width=0.85\columnwidth]{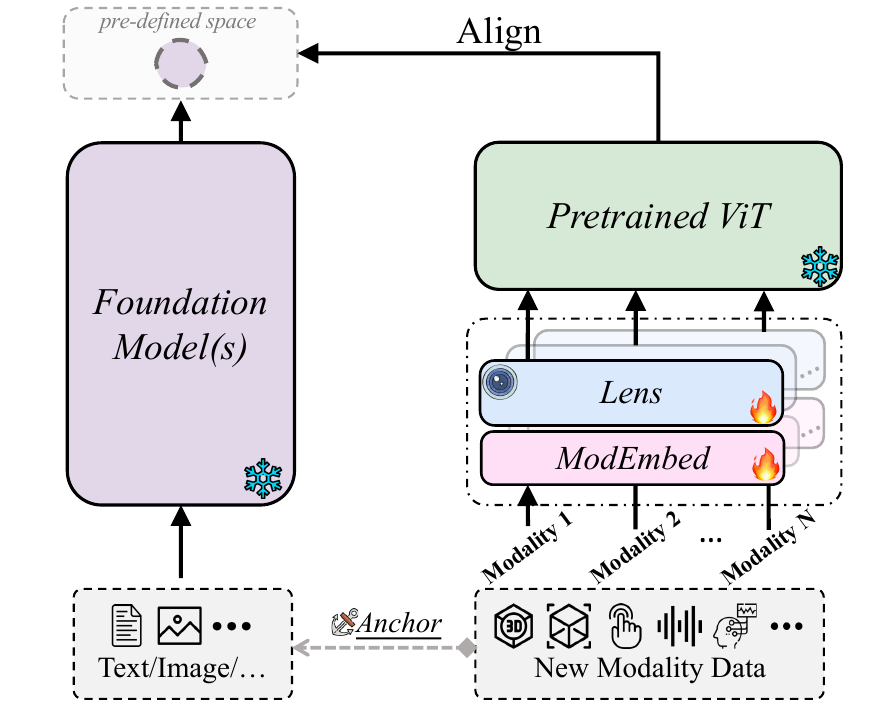}
  \caption{\textbf{Training Pipeline.} \methodname extends the capabilities of a \ptvit to diverse modalities. For each novel modality, it firstly employs a Modality Embedding (ModEmbed) and a Lens to learn mapping modality-specific data into an intermediate embedding space. 
  It subsequently employs a set of \ptvit layers to encode the feature.
  Finally, the output feature is aligned with the feature extracted from the anchor data (image, text, \emph{etc}.) of the new modality using an off-the-shelf foundation model.}
  \label{fig:method}
\vspace{-1.5em}
\end{figure}

\head{Overview.} \methodname advances omni-modal representation learning by leveraging \ptvit parameters to encode features for various modalities, utilizing the pre-existing knowledge in ViT.
Specifically, a modality-specific encoder, composed of a modality embedding module, the modality-specific Lens, and the \ptvit, is optimized to embed robust representations through the training objective of cross-modal alignment.
For each modality, we consider its associated anchor modalities, like image and text, as reference points for learning. For alignment, we leverage foundation models, such as CLIP~\cite{openai_clip}, to extract features from the anchor modalities.
Our approach leverages the extensive knowledge embedded in both the foundation models and \ptvit, providing a strong basis for representation learning for each modality. This compensates for the shortage of large-scale training data available for certain modalities.
We illustrate our approach in~\cref{fig:method}.


\subsection{Architecture}
\label{sec:arch}
\textbf{Foundation Models for alignment.}
In \methodname, the new modalities are aligned to a unified feature space established by a robust foundation model. Various options exist for this model, ranging from language models~\cite{bert, roberta, gpt1, gpt2, llama}, vision models~\cite{caron2021dino, oquab2023dinov2, he2022masked, fang2023eva} to vision-language models~\cite{openai_clip, cherti2022openclip, blip, blip2}. During training, we fix the foundation model's parameters and utilize it to encode features for the anchor data, which serves as supervision for feature alignment.

\par\head{Modality Encoder.} As is shown in~\cref{fig:method}, the modality encoder in \methodname consists of a Modality Embedding Module (ModEmbed), a Lens and a set of \ptvit layers.
Due to the distinct characteristics of various modalities, raw signals may not match the \ptvit input space. This mismatch can result in suboptimal performance, despite utilizing a powerful model.
Therefore, we employ some heuristic designs: 
(1) \textit{Obtain modality token embeddings:} for each modality, we adopt a specific tokenization scheme to transform raw input signals into token embeddings. 
(2) \textit{Map modality token embeddings to the \vit input space:} the Lens learns to map the modality embeddings into a group of latent embeddings, thereby constructing the input for the \ptvit.
Subsequently, the latent embeddings are forwarded to frozen \ptvit layers to obtain the final representation.

During training, the \ptvit component remains frozen, and only the parameters of ModEmbed and Lens are updated. More details can be found in Supp.
\begin{figure}[!h]
  \centering
  \includegraphics[width=0.87\columnwidth]{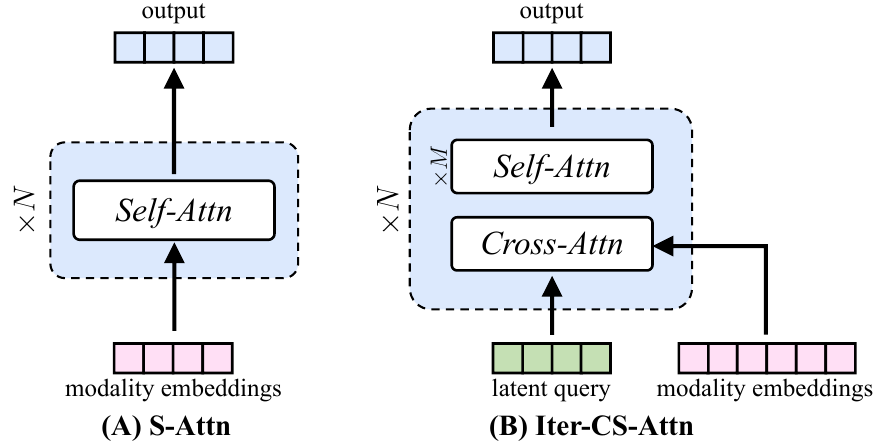}
  \vspace{-.5em}
  \caption{\textbf{Lens Architecture} used in \methodname.
}
  \label{fig:lens_variant}
  \vspace{-1.em}
\end{figure}

\par\head{Lens: Connecting Modalities to \vit.} We introduce two variants of Lens to link modality token embeddings to \vit. We show their architectures in~\cref{fig:lens_variant}.
\begin{itemize}[leftmargin=*]
\setlength{\itemsep}{2pt}
\setlength{\parsep}{0pt}
\setlength{\parskip}{0pt}
    \item \textbf{Self-attention blocks (S-Attn).} This variant involves a stack of self-attention layers~\cite{vaswani2017attention} that transforms the input token embeddings into intermediate embeddings with equal indices. We can potentially enhance this variant's capability by initializing it with pretrained weights from existing \vit layers. It suits modalities structured with image-like inputs, such as depth maps.
    \item \textbf{Iterative cross-self-attention blocks (Iter-CS-Attn).} This variant's basis block involves a cross-attention module coupled with a self-attention tower, inspired by~\cite{jaegle2021perceiver}. It maps a latent array and input embeddings to a latent embedding of matching length within the input latent array. This manner condenses inputs of varied sized into a latent bottleneck, 
    making it apt for lengthy input modalities like 3D point clouds.
    Similar architectures are employed in Vision-Language Models (VLMs)~\cite{alayrac2022flamingo,blip2} to extract visual information for Large Language Models (LLMs). Our innovation lies in utilizing this structure to map signals from diverse modalities into the \ptvit's input space, enabling the \vit to understand modalities beyond images.
\end{itemize}

\subsection{Training Objective}
\label{sec:mm_alignment}
In this work, we study modalities of 3D, depth, audio, tactile and EEG, among which all the data samples are associated with text descriptions, image appearances, or both. 
We use the pretrained CLIP~\cite{openai_clip,cherti2022openclip} as the foundation model. By default, we employ pretrained layers of \vit in the foundation CLIP as part of the modality encoder. Following the approach in previous works~\cite{xue2023ulip,xue2023ulip2,liu2023openshape,guzhov2022audioclip}, we adopt multimodal contrastive learning for representation alignment. 

We denote $X = \left\{x_1, \ldots, x_N \right\}$ as the collection of modality data to be learned, $\mathcal{A} = \left\{A_1, \ldots, A_M\right\}$ as the set of anchor modalities, $a_n^m$ as the anchor data of $x_n$ from modality $A_m$, $\ve{G}_A$ as the foundation model for anchor modality $A$, and $\ve{F}$ as the modality encoder to be learned. The contrastive loss for alignment is formulated as:
{\small
\begin{align*}
 \mathcal{L} = -\frac{1}{2B|\mathcal{A}|}\sum_{i=1}^B \sum_{k=1}^{|\mathcal{A}|} 
    \left(
         \log \frac{\exp(h_i^X \cdot h_i^{A_k} / \tau)}{\sum_j \exp (h_i^X \cdot h_j^{A_k} / \tau)} \right. \\ \left.
        + \log \frac{\exp( h_i^{A_k} \cdot h_i^X  / \tau)}{\sum_j \exp (h_i^{A_k} \cdot h_j^{X} / \tau)}
    \right),
\end{align*}
}where $B$ is the batch size; $\tau$ is a learnable temperature; $h_i^X = {\rm Norm}\left(\ve{F}(x_i)\right)$, $h_i^{A_k}={\rm Norm}\left(\ve{G}_{A_k}(a_i^k)\right)$ are normalized features of data $x_i$ and its anchor data $a_i^k$ from $A_k$.

\subsection{Free Lunch for Multimodal Foundation Models}
\begin{figure}[th]
  \centering
  \includegraphics[width=0.95\columnwidth]{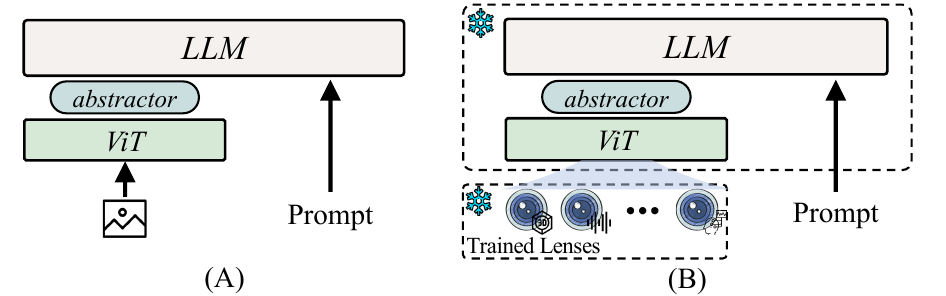}
  \vspace{-.5em}
  \caption{\textbf{Demonstration of integrating \methodname to MFM.} (A) Original overall pipeline of MFM for vision; (B) Illustration of plugging well-trained Lenses of different modalities to MFM, {\bf without} additional instruction-following training.
}
  \label{fig:mfm_plugin}
  \vspace{-.5em}
\end{figure}

\head{Plug \methodname into MFM.}
Recent MFMs for vision~\cite{zhu2022minigpt4,dai2023instructblip,liu2023llava,liu2023llava1.5,ge2023seed_tokenizer,ge2023seed_llama} enables LLMs to understand visual content. As shown in~\cref{fig:mfm_plugin}\textcolor{red}{A}, this process begins with the use of a frozen \vit to extract visual features. Subsequently, a well-trained abstractor module processes these features, constructing inputs that can be understood by the LLM.

By incorporating the \vit from MFM as part of the modality encoder and as the foundation model in \methodname training, the yielded modality Lenses can be seamlessly integrated into the MFM for plug-and-play application, as depicted in~\cref{fig:mfm_plugin}\textcolor{red}{B}. 
In later experiments, we showcase the emergent abilities facilitated by this tuning-free adaptation.

\section {Experiments}
\label{sec:experiments}
\subsection{Experimental Setup}\label{sec:exp_setup}
For this part, we describe the main experimental setup and provide full details in Supp.
\begin{table}[!h]
\centering
\resizebox{\linewidth}{!}{
\setlength{\tabcolsep}{2pt}
\begin{tabular}{l|cccc}
\bf Dataset             & \bf Task               & \bf \#cls & \bf Metric & \bf \#test \\ \Xhline{2\arrayrulewidth}
\threeDIcon{} ModelNet40(MN40)~\cite{wu2015modelnet}          & 3D shape cls       & 40    & Acc    & 2,468   \\
\threeDIcon{} Objaverse-LVIS(O-LVIS)~\cite{deitke2023objaverse}      & 3D shape cls       & 1,156  & Acc    & 46,832  \\
\threeDIcon{} ScanObjectNN(SONN)~\cite{uy2019scanobjectnn}        & 3D shape cls       & 15    & Acc    & 581   \\ \hline
\depthIcon{} SUN Depth-only(SUN-D)~\cite{song2015sun-rgbd}      & Scene cls          & 19    & Acc    & 4,660   \\
\depthIcon{} NYU-v2 Depth-only(NYU-D)~\cite{Silberman2012nyudepthv2}   & Scene cls          & 10    & Acc    & 654    \\ \hline
\audioIcon{} Audioset Audio-only(AS-A)~\cite{audioset} & Audio cls          & 527   & mAP    & 17,132\tablefootnote{\# test samples may differ from those used in previous work due to the unavailability of certain data.\label{fnref}} \\
\audioIcon{} ESC 5-folds(ESC)~\cite{piczak2015esc50}         & Audio cls          & 50    & Acc    & 2,000  \\
\audioIcon{} Clotho(Clotho)~\cite{drossos2020clotho}              & Retrieval          & -     & Recall & 1,046   \\
\audioIcon{} AudioCaps(ACaps)~\cite{audiocaps}           & Retrieval          & -     & Recall & 813\footref{fnref}    \\
\audioIcon{} VGGSound(VGGS)~\cite{chen2020vggsound}            & Audio cls          & 309   & Acc    & 15,434\footref{fnref}  \\ \hline
\tactileIcon{} Touch-and-go(TAG-M)~\cite{yang2022touch_and_go}        & Material cls       & 20    & Acc    & 29,879  \\
\tactileIcon{} Touch-and-go(TAG-H/S)~\cite{yang2022touch_and_go}        & Hard/Soft cls      & 2     & Acc    & 29,879  \\
\tactileIcon{} Touch-and-go(TAG-R/S)~\cite{yang2022touch_and_go}        & Rough/Smooth cls   & 2     & Acc    & 8,085   \\ \hline
\eegIcon{} ImageNet-EEG(IN-EEG)~\cite{spampinato2017eeg_data}                 & Visual Concept cls & 40    & Acc    & 1,997  
\end{tabular}
}
\vspace{-0.5em}
\caption{\textbf{Details of Downstream Datasets} across various modalities including 3D, depth, audio, tactile, and EEG. The evaluation is performed following feature alignment. The information presented includes the task type (classification/retrieval), the number of classes, the evaluation metric (Accuracy/mean Average Precision/Recall), and the quantity of test samples in each dataset.}
\label{tab:downstream_dataset}
\vspace{-0.5em}
\end{table}

\par\head{Pretraining Datasets.} 
Beyond image/video and text, we train \methodname on a variety of modalities, including 3D point cloud, depth, audio, tactile, and EEG data. These datasets are anchored to text descriptions, images, or both for feature alignment.

For 3D point cloud experiments, we utilize a combination of ShapeNet~\cite{chang2015shapenet}, 3D-FUTURE~\cite{fu2021-3d-future}, ABO~\cite{collins2022abo}, and Objaverse~\cite{deitke2023objaverse}.  We incorporate rendered images and text captions from previous works, resulting in three pretraining datasets: ULIP-ShapeNet~\cite{xue2023ulip}, ULIP-2-Objaverse~\cite{xue2023ulip2}, and OpenShape-Triplets~\cite{liu2023openshape}.
Depth data is sourced from the SUN RGB-D dataset~\cite{song2015sun-rgbd}, utilizing paired image and scene labels for alignment.
Audio data is obtained from the Audioset dataset~\cite{audioset}, accompanied by associated video and text label metadata.
Tactile data is sourced from the Touch-and-go dataset~\cite{yang2022touch_and_go}, featuring paired frame and material label text.
Finally, EEG data from~\cite{spampinato2017eeg_data} is aligned with paired ImageNet image and text labels.

\par\head{Evaluation on Downstream Understanding Tasks.}
We  evaluate \methodname across diverse modalities and protocols via a comprehensive set of downstream tasks. The primary datasets used for evaluation are summarized in~\cref{tab:downstream_dataset}.

\begin{table*}[t]
\centering
\begin{subtable}[h]{0.338\textwidth}
\centering
\resizebox{\textwidth}{!}{%
\setlength{\tabcolsep}{2pt}
\begin{tabular}{c|cc}
\threeDIcon{}               &  Top1 & Top5 \\ \Xhline{2.5\arrayrulewidth} 
\multicolumn{3}{l}{\textit{Trained on ULIP-ShapeNet}~\cite{xue2023ulip}} \\
ULIP-PointNet++(ssg)~\cite{xue2023ulip} &  55.7 & 75.7 \\
ULIP-PointNet++(msg)~\cite{xue2023ulip} &  58.4 & 78.2 \\
ULIP-PointMLP~\cite{xue2023ulip}        &  61.5 & 80.7 \\
ULIP-PointBERT~\cite{xue2023ulip}       &  60.4 & 84.0 \\
  \vitlensB       &   65.4 &  92.7 \\
  \vitlensL  &   \textbf{70.6} &  \textbf{94.4} \\ \hline
 \multicolumn{3}{l}{\textit{Trained on ULIP2-Objaverse}~\cite{xue2023ulip2}} \\
 ULIP2-PointNeXt~\cite{xue2023ulip2}     &  49.0 & 79.7 \\
ULIP2-PointBERT~\cite{xue2023ulip2}     &  70.2 & 87.0 \\
 \vitlensB  &  74.8 &  93.8 \\
 \vitlensL &  \textbf{80.6} &  \textbf{95.8} 
\end{tabular}
}
\caption{Zero-shot 3D of classification on ModelNet40. Models are pretrained on  triplets from ULIP-ShapeNet and ULIP2-Objaverse respectively.}\label{tab:3d_zero_shot_ulip}
\end{subtable}
\hfill
\begin{subtable}[h]{0.62\textwidth}
\centering
\resizebox{\textwidth}{!}{%
\setlength{\tabcolsep}{2pt}
\begin{tabular}{c|ccc|ccc|ccc}
\multirow{2}{*}{\threeDIcon{}} &  \multicolumn{3}{c|}{Objaverse-LVIS} & \multicolumn{3}{c|}{ModelNet40} & \multicolumn{3}{c}{ScanObjectNN} \\ 
                       & Top1       & Top3       & Top5      & Top1      & Top3     & Top5     & Top1      & Top3      & Top5     \\ \Xhline{2.5\arrayrulewidth}
\multicolumn{10}{l}{\textit{2D inference, no 3D training}} \\
PointCLIP~\cite{zhang2022pointclip}  & 1.9        & 4.1        & 5.8       & 19.3      & 28.6     & 34.8     & 10.5      & 20.8      & 30.6     \\
PointCLIP v2~\cite{zhu2022pointclipv2} & 4.7        & 9.5        & 12.9      & 63.6      & 77.9     & 85.0     & 42.1      & 63.3      & 74.5     \\ \hline
\multicolumn{10}{l}{\textit{Trained on OpenShape-Triplets (No LVIS)}~\cite{liu2023openshape}} \\
ULIP-PointBERT~\cite{xue2023ulip}   & 21.4       & 38.1       & 46.0      & 71.4      & 84.4     & 89.2     & 46.0      & 66.1      & 76.4     \\
OpenShape-SparseConv~\cite{liu2023openshape}   & 37.0       & 58.4       & 66.9      & 82.6      & 95.0     & 97.5     & 54.9      & 76.8      & 87.0     \\
OpenShape-PointBERT~\cite{liu2023openshape}   & 39.1       & 60.8       & 68.9      & 85.3      & 96.2     & 97.4     & 47.2      & 72.4      & 84.7     \\
 \vitlensG    &  \textbf{50.1}          &  \textbf{71.3}        &  \textbf{78.1}         &  \textbf{86.8}        &  \textbf{96.8}        &  \textbf{97.8}       &  \textbf{59.8}        &  \textbf{79.3}         &  \textbf{87.7}       \\ \hline
\multicolumn{10}{l}{\textit{Trained on OpenShape-Triplets}~\cite{liu2023openshape}} \\
ULIP-PointBERT~\cite{xue2023ulip}  & 26.8       & 44.8       & 52.6      & 75.1      & 88.1     & 93.2     & 51.6      & 72.5      & 82.3     \\
OpenShape-SparseConv~\cite{liu2023openshape}   & 43.4       & 64.8       & 72.4      & 83.4      & 95.6     & 97.8     & 56.7      & 78.9      & 88.6     \\
OpenShape-PointBERT~\cite{liu2023openshape}   & 46.8       & 69.1       & 77.0      & 84.4      & 96.5     & 98.0     & 52.2      & 79.7      & 88.7     \\
 \vitlensG  &  \textbf{52.0}   &  \textbf{73.3}   &  \textbf{79.9}     &  \textbf{87.6}     &  \textbf{96.6}    &  \textbf{98.4}     &  \textbf{60.1}      &  \textbf{81.0}     &  \textbf{90.3}  \\ 
\end{tabular}}

\caption{Zero-shot 3D classification on Objaverse-LVIS, ModelNet40 and ScanObjectNN. Models are pretrained on OpenShape Triplets. ``NO LVIS'' denotes exclude the Objaverse-LVIS subset.}\label{tab:3d_zero_shot_openshape}
\end{subtable}
\vspace{-0.5em}
\caption{Zero-shot 3D classification on downstream datasets,  
measured in accuracy(\%).} \label{tab:3d_zeroshot_all}
\vspace{-1em}
\end{table*}

\par\head{Main Implementation Details.}
We use the pretrained vision and text encoders from OpenCLIP~\cite{cherti2022openclip}. We apply different model sizes: \vitlensB based on \vit-B/16, \vitlensL based on \vit-L/14, and \vitlensG based on \vit-bigG/14.

For 3D point cloud data, we follow the baseline methods~\cite{xue2023ulip,liu2023openshape} to uniformly sample 8,192 or 10,000 points and grouping them into sub-clouds through Farthest Point Sampling (FPS) followed by KNN grouping of neighboring points.
For depth input, we follow~\cite{girdhar2023imagebind} to use in-filled depth values and convert them to disparity for scale normalization.
For audio data, we sample 5-second clips and extract a single frame randomly from the video clip if video serves as anchor data. The audio waveform is converted into a sequence of 128-dimensional log Mel filterbank features using a 25ms Hamming window every 10ms, following~\cite{gong21bast}.
For tactile input, we use RGB data collected from GelSight~\cite{johnson2009gelsight}.
For EEG signals, we employ 128-channel temporal sequences and use the frequency range of 5-95Hz following~\cite{bai2023dreamdiffusion}.

\subsection{Results on Understanding Tasks}
\head{Zero-shot 3D Classification.}
We follow~\cite{xue2023ulip,xue2023ulip2,liu2023openshape} to use (point cloud, image, text) triplets to train \methodname. We conduct zero-shot classification on downstream benchmarks. The overall results can be found in~\cref{tab:3d_zeroshot_all}.
In particular, when pretrained on ULIP-ShapeNet or ULIP2-Objaverse, \methodname outperforms ULIP with different 3D encoders~\cite{qi2017pointnet++,ma2022pointmlp,qian2022pointnext,yu2022pointbert}, as is shown in~\cref{tab:3d_zero_shot_ulip}.

We present the results of training on OpenShape-Triplets in~\cref{tab:3d_zero_shot_openshape}. To align with~\citep{liu2023openshape}, we adopt \vitlensG and train on both ``NO LVIS'' (excluding all shapes from the Objaverse-LVIS subset) and the entire set for comparison. \methodname outperforms models adopted in OpenShape~\cite{liu2023openshape}.
Notably, \methodname significantly improves the accuracy on the long-tail categories of Objaverse-LVIS, from 46.8\% to 52.0\%. Additionally, when trained on the NO LVIS subset, \methodname achieves a top-1 accuracy of 50.1\%. This performance beats ULIP by roughly 30\% and surpasses OpenShape-PointBERT trained on the entire set by 3.3\%, demonstrating the data-efficient merit of \methodname.
Regarding ModelNet40, \methodname achieves an 87.4\% accuracy, surpassing previous SOTA. Moreover, on ScannetObjectNN, containing challenging real scans with noise and occlusion, our method exhibits decent sim-to-real transfer ability. It achieves a 60.1\% zero-shot accuracy without specific sim-to-real training, surpassing the previous SOTA.

\begin{table*}[h]
\centering
  \begin{minipage}{.63\textwidth} 
    \centering
        \resizebox{\textwidth}{!}{%
        \setlength{\tabcolsep}{4pt}
        \begin{tabular}{c|c|c|c|c|cc|cc}
        \multirow{2}{*}{\audioIcon{}} & \multirow{2}{*}{\textit{anchor}} & AudioSet & VGGSound$^\diamond$  & ESC$^\diamond$  & \multicolumn{2}{c|}{Clotho$^\diamond$} & \multicolumn{2}{c}{AudioCaps$^\diamond$} \\
                                &                         & mAP      & Top1      & Top1 & R@1          & R@10         & R@1           & R@10          \\ \Xhline{3\arrayrulewidth}
        \multicolumn{1}{l|}{AVFIC~\cite{nagrani2022avfic}}                 & -                       & -        & -         & -    & 3.0          & 17.5         & 8.7           & 37.7          \\
        \multicolumn{1}{l|}{ImageBind-H~\cite{girdhar2023imagebind}}          & I                       & 17.6     & 27.8      & 66.9 & 6.0          & 28.4         & 9.3           & 42.3          \\
        \multicolumn{1}{l|}{\vitlensL}                & I                       & 23.1     & 28.2      & 69.2 & 6.8          & 29.6         & 12.2          & 48.7          \\
        \multicolumn{1}{l|}{AudioCLIP~\cite{guzhov2022audioclip}}               & I+T                     & 25.9     & -         & 69.4 & -            & -            & -             & -             \\
        \multicolumn{1}{l|}{\vitlensL}                & I+T                     & \bf 26.7     & \bf 31.7      & \bf 75.9 & \bf 8.1          & \bf 31.2         & \bf 14.4          & \bf 54.9          \\ \hline
        \multicolumn{1}{l|}{Prev. ZS SOTA}           & -                       & -        & \color{DarkGray}29.1/46.2$^\star$~\cite{laionclap2023} & \color{DarkGray}91.8~\cite{wang2023onepeace} & 6.0          & 28.4~\cite{girdhar2023imagebind}         & 9.3       & 42.3~\cite{girdhar2023imagebind}   
        \end{tabular}
        }
        \vspace{-0.5em}
        \caption{Audio classification and retrieval on Audioset, VGGSound, ESC, Clotho and AudioCaps. $^\diamond$denotes zero-shot evaluation. {\color{DarkGray} Gray-out} denotes using larger audio-text datasets in pretraining. {\color{DarkGray}$^\star$}denotes using augmented captions for training.}\label{tab:audio_cls}
  \end{minipage}%
  \hfill
  \begin{minipage}{.345\textwidth}
    \centering
    \resizebox{\textwidth}{!}{
    \setlength{\tabcolsep}{2pt}
    \begin{tabular}{l|c|ccc}
    \multicolumn{1}{c|}{\videoIcon{} \audioIcon{}}         & \textit{modality} & R@1  & R@5  & R@10 \\ \Xhline{3\arrayrulewidth}
    \color{DarkGray} MIL-NCE~\cite{miech2019howto100m} & \color{DarkGray} V & \color{DarkGray} 8.6 & \color{DarkGray} 16.9 & \color{DarkGray} 25.8 \\
    \color{DarkGray} SupportSet~\cite{patrick2021supportset} & \color{DarkGray} V & \color{DarkGray} 10.4 & \color{DarkGray} 22.2 & \color{DarkGray} 30.0 \\
    \color{DarkGray} AVFIC~\cite{nagrani2022avfic}          & \color{DarkGray} A+V      & \color{DarkGray} 19.4 & \color{DarkGray} 39.5 & \color{DarkGray} 50.3 \\
    ImageBind-H~\cite{girdhar2023imagebind}    & A+V      & 36.8 & 61.8 & 70.0 \\
    \vitlensL       & A+V      & \bf 37.6 & \bf 63.2 & \bf 72.6 \\ \hline
    \color{DarkGray}Zero-shot SOTA~\cite{chen2023vast} & \color{DarkGray}V        & \color{DarkGray}49.3 & \color{DarkGray}68.3 & \color{DarkGray}73.9
    \end{tabular}
    }
    \vspace{-0.5em}
    \caption{Video Retrieval on MSRVTT. V: use video; A+V: use audio and video. {\color{DarkGray}Gray-out} means using video data in pretraining.}\label{tab:video_ret}
  \end{minipage}
    \begin{minipage}{.33\textwidth} 
    \centering
    \resizebox{\textwidth}{!}{
        \setlength{\tabcolsep}{2pt}
        \begin{tabular}{l|c|cc}
        \multicolumn{1}{c|}{\depthIcon{}}      & \textit{anchor} & NYU-D & SUN-D \\ \Xhline{3\arrayrulewidth}
        Text Paired~\cite{girdhar2023imagebind} & T$^\star$     & 41.9  & 25.4  \\
        ImageBind-H~\cite{girdhar2023imagebind} & I      & 54.0  & 35.1  \\
        \vitlensL    & I      & 64.2  & 37.4  \\
        \vitlensL    & I+T     & \bf 68.5  & \bf 52.2  \\ \hline
        \color{DarkGray} Supervised SOTA~\cite{girdhar2022omnivore} &  
        \color{DarkGray}-   & \color{DarkGray} 76.7  & \color{DarkGray} 64.9
        \end{tabular}
    }
        \vspace{-0.5em}
        \caption{Depth-only scene classification on NYU-D and SUN-D. $^\star$\cite{girdhar2023imagebind} rendered depth as grayscale images for direct testing. The supervised SOTA~\cite{girdhar2022omnivore} used RGBD as input and extra training data.}\label{tab:depth_cls}
  \end{minipage}%
  \hfill
  \begin{minipage}{.315\textwidth}
    \centering
    \resizebox{\textwidth}{!}{
        \setlength{\tabcolsep}{2pt}
        \begin{tabular}{l|c|ccc}
        \multicolumn{1}{c|}{\tactileIcon{}}                   & \textit{anchor} & Material & H/S & R/S \\ \Xhline{3\arrayrulewidth}
        ImageBind-B$^\ast$             & I      & 24.2     & 65.7      & 69.8                             \\
        \vitlensB                 & I      & 29.9     & 72.4   & 77.9                           \\
        \vitlensL                 & I      &  31.2        &    74.3       &         \bf 78.2                        \\
        \vitlensL                 & I+T     &  \bf 65.8       &    \bf 74.7       &          63.8                        \\ \hline
        \multicolumn{5}{l}{\textit{Linear Probing}} \\
        CMC~\cite{tian2020cmc,yang2022touch_and_go}                      & I      & 54.7     & 77.3      & 79.4                             \\
        \vitlensB  & I      & \bf 63.0     & \bf 92.0      & \bf 85.1               
        \end{tabular}
    }
        \vspace{-0.5em}
        \caption{Tactile classification on Touch-and-go. $^\ast$denotes our implementation. H/S: Hard/Soft; R/S: Rough/Smooth.}\label{tab:tactile_cls}
  \end{minipage}%
  \hfill
  \begin{minipage}{.32\textwidth}
    \centering
    \resizebox{\textwidth}{!}{
        \setlength{\tabcolsep}{3pt}
        \begin{tabular}{l|c|cc}
        \multicolumn{1}{c|}{\eegIcon{}}           & \textit{anchor} & Val  & Test \\ \Xhline{3\arrayrulewidth}
        ImageBind-B$^\ast$      & I      & 17.3 & 18.4 \\
        DreamDiffusion-L\textsuperscript{\#}~\cite{bai2023dreamdiffusion} & I      & 20.4 & 19.2 \\
        \vitlensB         & I      & 24.6 & 25.3 \\
        \vitlensL         & I      &  29.3    &   29.2   \\
        \vitlensL         & I+T     & \bf 41.8 & \bf 42.7
        \end{tabular}
    }
\vspace{-0.5em}
\caption{Visual concept classification on ImageNet-EEG. $^\ast$denotes our implementation. \textsuperscript{\#}We use the released EEG encoder and paired text encoder for inference. We report results on Val and Test set.}\label{tab:eeg_cls}
  \end{minipage}%
\vspace{-.5em}
\end{table*}

\head{Audio Classification and Retrieval.}
In our comparison presented in~\cref{tab:audio_cls}, \vitlensL consistently outperforms prior approaches in both audio classification and text-to-audio retrieval tasks.  When aligned to images (I), \vitlensL outperforms ImageBind based on Huge CLIP~\cite{cherti2022openclip}, and AVFIC~\cite{nagrani2022avfic}, which leverages automatically mined audio-text pairs for alignment.
When aligned to images and texts (I+T), \vitlensL demonstrates stronger performance and significantly outperforms AudioCLIP~\cite{guzhov2022audioclip}. Although AudioCLIP uses a audio encoder pretrained with Audioset supervised classification, it falls behind \vitlensL.
Additionally, on zero-shot VGGSound classification, \vitlensL surpasses the SOTA~\cite{laionclap2023} when class names are used as text supervision for alignment.

\head{Audio and Video Retrieval.}
We use the MSR-VTT~\cite{Xu2016msrvtt} benchmark to evaluate the text to audio and video retrieval performance, as presented in~\cref{tab:video_ret}. We follow~\cite{girdhar2023imagebind} to combine audio (A) and video (V) modalities. \methodname outperforms several prior methods, even surpassing those that incorporate video data for training~\cite{miech2019howto100m,patrick2021supportset,nagrani2022avfic}.

\head{Depth-only Scene Classification.}
In~\cref{tab:depth_cls}, we present our results for depth-only classifications. \methodname outperforms ImageBind across SUN-D and NYU-D. By using image and text as anchor data, \methodname further improves the performance and narrows the gap with the supervised SOTA model~\cite{girdhar2022omnivore} with extra training data.

\head{Tactile Classification Tasks.}
Results for tactile tasks are displayed in~\cref{tab:tactile_cls}. Across various tactile classification tasks like material, hard/soft, and rough/smooth classification, \vitlensB demonstrates superior performance compared to our implementation of ImageBind-B.  
Even trained with appearance or text labels for material, \methodname can perform well on the hard/soft and rough/smooth classification tasks. This underscores the extensive knowledge transfer by CLIP during training.
Furthermore, scaling up to a larger model and incorporating text during training can further boost the performance.
In comparing the image-aligned \vitlensB with CMC~\cite{yang2022touch_and_go} using linear probing, we observe significantly superior performance by \methodname.

\head{EEG Visual Concept Classification.} 
Results in~\cref{tab:eeg_cls} show that \methodname consistently outperforms our implemented ImageBind-B. Additionally, when compared to the EEG encoder from~\cite{bai2023dreamdiffusion}, which used more EEG data for MAE-style pretraining~\cite{he2022masked} and then aligned with the CLIP-L14 image encoder, \methodname demonstrates superior performance.

\subsection{Few-shot Linear Probing}
\begin{figure}[h!]
    \resizebox{0.49\linewidth}{!}{
        \begin{tikzpicture}
    \begin{axis}[
        xtick={0, 1, 2, 4, 8}, %
        legend pos=south east,
        xticklabels={0, 1, 2, 4, 8}, %
        xmin=0,
        ymin=-5,
        grid=both,
        grid style={line width=.1pt, draw=gray!10},
        major grid style={line width=.2pt,draw=gray!30},
        minor tick num=2,
        axis x line*=bottom,
        axis y line*=left,
        width=\linewidth,
        ylabel style= {align=center, font=\large},
        xlabel style = {font=\large},
        ylabel={\depthIcon{} SUN-D Top-1},
        ylabel near ticks,
        xlabel={\# of labeled training samples per class},
        yticklabel style = {font=\large},
        xticklabel style = {font=\large},
        legend style={cells={align=left}, font=\large, fill=none, draw=none},
    ]
    
    \addplot[mark=square, very thick, pearThree, mark options={solid}, line width=2.5pt, mark size=3pt] plot coordinates {
        (1, 26.6) %
        (2, 36.8) %
        (4, 40.2) %
        (8, 43.3) %
    };
    \addlegendentry{\vitlensL$\rightarrow$ I}

    \addplot[mark=o, very thick, gray!60, mark options={solid}, line width=2.5pt, mark size=3pt] plot coordinates {
        (1, 20.2) %
        (2, 32.8) %
        (4, 34.8) %
        (8, 36.9) %
    };
    \addlegendentry{ImageBind-H~\cite{girdhar2023imagebind}}

    \addplot[mark=triangle, very thick, gray!60, mark options={solid}, line width=2.5pt, mark size=3pt %
    ] plot coordinates {
        (1, 8.6) %
        (2, 14.2) %
        (4, 16.0) %
        (8, 20.6) %
    };
    \addlegendentry{MultiMAE~\cite{bachmann2022multimae}}

    \addplot[mark=o, very thick, gray!60, mark options={mark size=3.2pt, solid, line width=2pt}] plot coordinates {
        (0, 35.1) %
    };

    \addplot[mark=square, very thick, pearThree, mark options={mark size=3.2pt, solid, line width=2pt}] plot coordinates {
        (0, 37.4) %
    };

    \end{axis}
\end{tikzpicture}
    } \hfill
    \resizebox{0.49\linewidth}{!}{
        \begin{tikzpicture}
    \begin{axis}[
        xtick={0, 1, 2, 4, 8}, %
        legend pos=south east,
        xticklabels={0, 1, 2, 4, 8}, %
        xmin=0,
        ymin=-5,
        grid=both,
        grid style={line width=.1pt, draw=gray!10},
        major grid style={line width=.2pt,draw=gray!30},
        minor tick num=2,
        axis x line*=bottom,
        axis y line*=left,
        width=\linewidth,
        ylabel style= {align=center, font=\large},
        xlabel style = {font=\large},
        ylabel={\threeDIcon{} O-LVIS Top-1},
        ylabel near ticks,
        xlabel={\# of labeled training samples per class},
        yticklabel style = {font=\large},
        xticklabel style = {font=\large},
        legend style={cells={align=left}, font=\large, fill=none, draw=none},
    ]
    
    \addplot[mark=square, very thick, pearDark, mark options={solid}, line width=2.5pt, mark size=3pt] plot coordinates {
        (1, 46.1) %
        (2, 53.8) %
        (4, 63.0) %
        (8, 73.4) %
    };
    \addlegendentry{\vitlensG}

    \addplot[mark=triangle, very thick, gray!60, mark options={solid}, line width=2.5pt, mark size=3pt] plot coordinates {
        (1, 21.2) %
        (2, 25.4) %
        (4, 31.8) %
        (8, 37.6) %
    };
    \addlegendentry{OS-SparseConv~\cite{liu2023openshape}}

    \addplot[mark=o, very thick, gray!60, mark options={solid}, line width=2.5pt, mark size=3pt] plot coordinates {
        (1, 23.9) %
        (2, 30.3) %
        (4, 36.1) %
        (8, 42.8) %
    };
    \addlegendentry{OS-PointBERT~\cite{liu2023openshape}}

    \addplot[mark=triangle, very thick, gray!60, mark options={mark size=3.2pt, solid, line width=2pt}] plot coordinates {
        (0, 37.0) %
    };

    \addplot[mark=o, very thick, gray!60, mark options={mark size=3.2pt, solid, line width=2pt}] plot coordinates {
        (0, 39.1) %
    };

    \addplot[mark=square, very thick, pearDark, mark options={mark size=3.2pt, solid, line width=2pt}] plot coordinates {
        (0, 52.0) %
    };

    \end{axis}
\end{tikzpicture}
    }
    \vspace{-0.5em}
    \caption{\textbf{Few-shot linear probing on depth and 3D point cloud.}
    We mark the zero-shot classification performance on the y-axis.
    We train linear classifiers on fixed features for the $\ge\!1$-shot settings.
    }
    \label{fig:few_shot_lp_cls}
    \vspace{-.8em}
\end{figure}
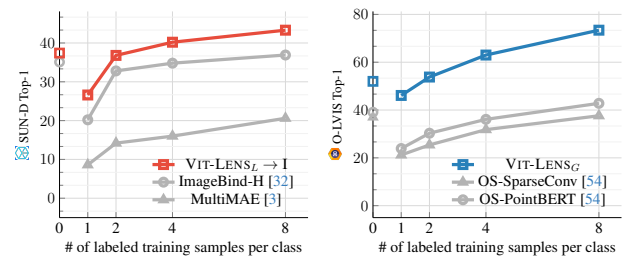
We evaluate the label-efficiency of \methodname through few-shot linear probing using \vitlensL(I) depth encoder and \vitlensG 3D encoder, as shown in~\cref{fig:few_shot_lp_cls}. Linear classifiers are trained on fixed representation features.
For few-shot depth linear probing, we compare with ImageBind~\cite{girdhar2023imagebind} and MultiMAE~\cite{bachmann2022multimae} trained on images, depth, and semantic segmentation data. \methodname consistently outperforms both methods in zero-shot and few-shot settings.
For few-shot 3D classification, we compare with SparseConv and PointBERT trained in~\cite{liu2023openshape}. \methodname significantly outperforms all methods by a large margin in all few-shot settings, showcasing its robust generalization capabilities.

\subsection{Results on \Boldmethodname MFMs}
\begin{figure*}[t]
    \centering
    \includegraphics[width=\linewidth]{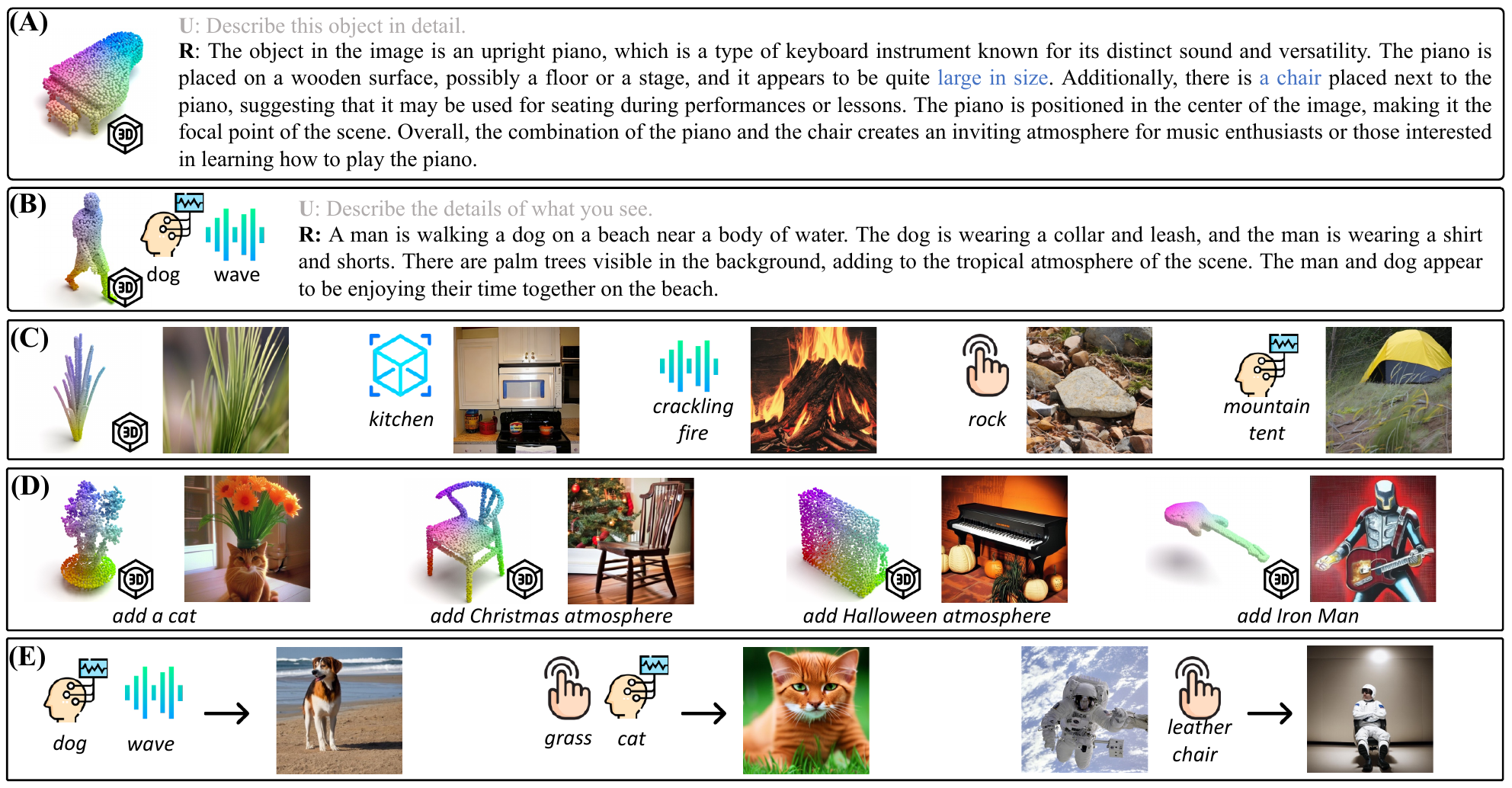}
    \vspace{-1.5em}
    \caption{
    \textbf{Qualitative examples for plugging \methodname into MFMs.}
    {\bf (A-B) Integrate with InstructBLIP:} Accurately capturing concepts from single (A) or multiple modalities (B), providing detailed descriptions based on InstructBLIP's instruction-following capability.
    {\bf (C-E) Integrate with SEED:} Extending SEED's capability to emergent compositional Any-to-image generation. (C) Single modality to image generation. (D) Text-guided any-to-image generation. (E) Multi-modalities-to-image generation.
    }
    \label{fig:vitlens_mfm}
    \vspace{-.5em}
\end{figure*}
In this section, we plug \methodname across various modalities into off-the-shelf MFMs, and show in our experimental results that the MFMs' capabilities can be transferred to novel modalities and their combinations, without instruction-following training.

\par \head{MFM Selection in Practice.}
In this work, we select InstructBLIP~\cite{dai2023instructblip} and SEED~\cite{ge2023seed_tokenizer,ge2023seed_llama} to probe the emergent capabilities of the MFMs with our \methodname plugged in. Both InstructBLIP and SEED utilize EVA01-g14 \clipvit~\cite{fang2023eva} as the visual encoder. Following the practice in~\cref{sec:mm_alignment}, we use the same \ptvit for \methodname training in MFMs experiments. More details can be found in Supp.

\head{InstructBLIP with \methodname.} 
InstructBLIP~\cite{dai2023instructblip} introduced a framework for instruction tuning in a vision-language model, demonstrating its capabilities in tasks like complex visual reasoning and image descriptions.  We show in our experiment that these capabilities can be effectively extended to novel modalities through the integration of \methodname.
Qualitative examples in~\cref{fig:vitlens_mfm} (A-B) showcase the model's ability to follow instructions across various modalities, enabling Any-modality QA, captioning, \emph{etc}.
Additionally, the model demonstrates precise and detailed descriptions, such as identifying a small ``chair'' next to a giant piano in (A), emphasizing the superior alignment achieved by \methodname.

\head{SEED with \methodname.}
SEED-LLaMa~\cite{ge2023seed_llama} is an MFM distinguished by its capacity for multimodal comprehension and image generation. This is achieved through multimodal pretraining and instruction tuning along with its SEED tokenizer~\cite{ge2023seed_tokenizer}.
We present qualitative results of integrating \methodname with SEED in~\cref{fig:vitlens_mfm} (E-G). The outcomes illustrate how the combined model extends SEED's capabilities to diverse modalities.
Examples in (E-G) show the ability of compositional any-to-image generation~\cite{ge2023seed_llama}. 
It can translate input from any modality into an image, generate an image based on a text prompt given input from any modality, and seamlessly blend visual concepts from combinations of any modalities into a coherent and plausible image.


\subsection{Ablation Study}
We conduct ablations studies to investigate the effectiveness of various designs for \methodname in omni-modal learning. We report the main results here and full details are in Supp.

\par\head{Lens designs for different modalities.}
We study the effect of Lens designs as outlined in~\cref{sec:arch} for different modalities. We use \vitlensB and set comparable amount of trainable parameters for the two variants. We also examine the effectiveness of initializing S-Attn type Lens with pretrained weights.
We train 3D point cloud on ULIP-ShapeNet and follow the main settings for other modalities.
The results are shown in~\cref{tab:abla_lens_design}.
We observe consistent performance enhancement by initializing S-Attn Lens with pretrained weights.  
For image-like inputs such as depth maps and RGB-based tactile data, the S-Attn design exhibits superiority.
 Conversely, modalities significantly different from image inputs, like 3D point clouds, audio spectrograms, and EEG, benefit more from Iter-CS-Attn design. Additionally, it reduces the computational overhead by reducing the input length for \vit. Further details are available in the Supp.
\begin{table}[t]
\centering
\resizebox{0.99\linewidth}{!}{%
\setlength{\tabcolsep}{2pt}
\begin{tabular}{c|ccccc}
 Test Dataset $\blacktriangleright$  & \threeDIcon{}MN40 & \depthIcon{}SUN-D & \audioIcon{}ESC & \tactileIcon{}TAG-M & \eegIcon{}IN-EEG \\ \Xhline{3\arrayrulewidth}
S-Attn w/o pt weights & 63.8    & 48.6     & 70.1   & 61.8   & 25.4      \\
S-Attn w/ pt weights  & 65.4    & \cellcolor{lavenderweb}50.9     & 70.9   & \cellcolor{lavenderweb}63.6   & 26.3      \\
Iter-CS-Attn & \cellcolor{lavenderweb}65.4    & 47.5     & \cellcolor{lavenderweb}71.2   & 60.6   & \cellcolor{lavenderweb}35.9      
\end{tabular}
}
\vspace{-0.5em}
\caption{{\bf Lens designs} for different modalities. All modalities are aligned to ``I+T''. Lens w/ pt weights means tuning corresponding Self-Attn blocks in the \ptvit, and w/o means random initialization. Default setting is marked with \colorbox{lavenderweb}{color box}.}\label{tab:abla_lens_design}
\vspace{-1.5em}
\end{table}

\par\head{Modality encoder designs and settings.}  We investigate the efficacy of integrating a set of \ptvit layers into the modality encoder. We use the same datasets for training and testing as in the Lens design ablation.
We compare \vitlensB with an architecture that combines ModEmbed and \vit and employing different settings for the \vit component, as detailed in~\cref{tab:abla_encoder_design}.
Results indicate that simply adding the ModEmbed to a \ptvit cannot fully exploit the the potential of the \ptvit (\#2). Training the entire encoder with pretrained weights outperforms training from scratch, highlighting the effectiveness of utilizing the pretrained weights for learning (\#1 vs \#3). 
In comparison to \#3, \vitlensB achieves comparable or better performance, especially for the less common modalities. 
Moreover, our \methodname employs fewer trainable parameters than training the entire encoder and reduces computational overhead for modalities with lengthy inputs. Consequently, by introducing Lens, \methodname effectively and efficiently transfers the capabilities of \ptvit to various modalities.
\begin{table}[h]
\vspace{-0.5em}
\centering
\resizebox{0.99\linewidth}{!}{%
\setlength{\tabcolsep}{2pt}
\begin{tabular}{c|ccccc}
 Test Dataset $\blacktriangleright$  & \threeDIcon{}MN40 & \depthIcon{}SUN-D & \audioIcon{}ESC & \tactileIcon{}TAG-M & \eegIcon{}IN-EEG \\ \Xhline{3\arrayrulewidth}
\multicolumn{1}{l|}{\#1 M.E. $\to$ \vit (scratch)}      & 62.4    & 46.3     & 68.8   & 55.6   & 20.5      \\
\multicolumn{1}{l|}{\#2 M.E. $\to$ \vit (pt, frozen)}   & 50.0    & 36.8     & 54.9   & 24.8   & 14.2      \\
\multicolumn{1}{l|}{\#3 M.E. $\to$ \vit (pt, tune)}     & 67.4    & 48.2     & 71.6   & 59.4   & 27.2      \\
\rowcolor{lavenderweb} \vitlensB & 65.4    & 50.9     & 71.2   & 63.6   & 35.9
\end{tabular}
}
\vspace{-0.5em}
\caption{{\bf Encoder designs and settings} for different modalities. All modalities are aligned to ``I+T''. M.E. denotes ModEmbed. \vitlensB is the default setting.}\label{tab:abla_encoder_design}
\vspace{-1.5em}
\end{table}

\par \head{Scaling up foundation model and \methodname.} We explore the effectiveness of scaling up \methodname for feature alignment. We conduct experiments to pretrain for 3D on ULIP-ShapeNet, and depth on SUN-D. 
While previous works~\cite{girdhar2023imagebind,liu2023openshape} show that scaling to a large encoder($>$100M) degrades the performance, we show in~\cref{fig:scale_model_size} that scaling up \methodname can improve the 3D and depth representation and enhance performance.
\begin{figure}[h!]
\vspace{-0.5em}
    \begin{subfigure}[b]{0.49\linewidth}
            \begin{tikzpicture}
    \begin{axis}[
        xtick={4.45, 5.71, 7.52}, 
        xticklabels={B, L, bigG},
        grid=both,
        grid style={line width=.1pt, draw=gray!10},
        major grid style={line width=.2pt,draw=gray!50},
        minor tick num=2,
        axis x line*=bottom,
        axis y line*=left,
        height=1.4in,
        width=\linewidth,
        ylabel style= {align=center},
        ylabel={\depthIcon{}SUN-D~\ref{pgf:depth_model_scaling_sun:img}},
        ylabel near ticks,
        yticklabel style = {font=\small},
        xticklabel style = {font=\small},
        legend style={cells={align=left}, font=\footnotesize},
    ]
    \addplot[mark=square, very thick, pearThree] plot coordinates {
        (7.52, 54.5) %
        (5.71, 52.2) %
        (4.45,  50.9) %
    };\label{pgf:depth_model_scaling_sun:img}
    \end{axis}
\end{tikzpicture}
    \end{subfigure}
    \hfill
    \begin{subfigure}[b]{0.49\linewidth}
            \begin{tikzpicture}
    \begin{axis}[
        xtick={4.45, 5.71, 7.52},
        xticklabels={B, L, bigG}, 
        grid=both,
        grid style={line width=.1pt, draw=gray!10},
        major grid style={line width=.2pt,draw=gray!50},
        minor tick num=2,
        axis x line*=bottom,
        axis y line*=left,
        height=1.4in,
        width=\linewidth,
        ylabel style= {align=center},
        ylabel={\threeDIcon{}MN40~\ref{pgf:3d_model_scaling_mn:img}},
        ylabel near ticks,
        yticklabel style = {font=\small},
        xticklabel style = {font=\small},
        legend style={cells={align=left}, font=\footnotesize},
    ]
    \addplot[mark=square, very thick, pearDark] plot coordinates {
        (7.52, 72.2) %
        (5.71, 70.6) %
        (4.45,  65.4) %
    };\label{pgf:3d_model_scaling_mn:img}
    \end{axis}
\end{tikzpicture}
    \end{subfigure}
    \vspace{-0.5em}
    \caption{\textbf{Scaling the \methodname on depth and 3D point cloud.} B: \vitlensB, L: \vitlensL, bigG: \vitlensG.}\label{fig:scale_model_size}
    \vspace{-0.5em}
\end{figure}

\par\head{Different pretrained-ViTs for \Boldmethodname.} We evaluate different \ptvit variants for omni-modal representation learning. We use CLIP-ViT-bigG/14 as the teacher foundation model and apply different ViTs for the modalitiy encoder. We use the same datasets for training and testing as in the model scaling ablation. Results in~\cref{tab:abla_vit_variant} demonstrate that the use of \ptvits including the self-supervised and CLIP pretrained variants, outperforms training from scratch on both depth and 3D modalities. This indicates that different \ptvits possess the potential to serve as effective omni-modal learners.
\begin{table}[!h]
\resizebox{0.99\linewidth}{!}{%
\setlength{\tabcolsep}{2pt}
    \begin{tabular}{l|ccccc}
    \begin{tabular}[c]{@{}c@{}}ViT\\ variant\end{tabular} $\blacktriangleright$ & \cellcolor{Gray}\begin{tabular}[c]{@{}c@{}}RndInit\\ ViT-B16\end{tabular}  &\begin{tabular}[c]{@{}c@{}}DINO~\cite{caron2021dino}\\ ViT-B16\end{tabular} & \begin{tabular}[c]{@{}c@{}}OpenCLIP\\ ViT-B16\end{tabular} & \begin{tabular}[c]{@{}c@{}}OpenCLIP\\ ViT-L14\end{tabular} & \begin{tabular}[c]{@{}c@{}}OpenCLIP\\ ViT-bigG14\end{tabular} \\ \Xhline{3\arrayrulewidth}
    \depthIcon{} SUN-D   &  \cellcolor{Gray}48.0  & 50.9    & 51.4   & 53.2  & 54.5  \\
    \threeDIcon{} M40N    &  \cellcolor{Gray}66.2  & 68.5    & 68.3   & 71.4  & 72.2                                                         
    \end{tabular}
}
\vspace{-0.5em}
\caption{{\bf Different \vit} for modality encoders in \methodname. we train the entire encoder for the baseline \colorbox{Gray}{RndInit} (random initialization) , while others follow \methodname training setting.}\label{tab:abla_vit_variant}
\vspace{-1em}
\end{table}

\section{Conclusion}
In this paper, we introduce \methodname, a straightforward yet effective method to advance omni-modal representations. 
\methodname employs a \ptvit to encode features for diverse modalities, eliminating the need for separate modality-specific architectures. 
The rich knowledge from large-scale data within the \ptvit also reduces the burden of extensive data collection. 
We train \methodname for various modalities, including 3D point cloud, depth, audio, tactile, and EEG. Experimental results in understanding tasks demonstrate that \methodname consistently achieves leading performance.
Moreover, we integrate \methodname to off-the-shelf MFMs, \emph{i.e.,} InstructBLIP and SEED, unlocking emergent capabilities such as any-modality instruction following, any-modality-to-image generation and text-guided any-modality-to-image editing.
Finally, we believe that \methodname will stimulate further research and innovation in the field of omni-modal representation learning, paving the way for more versatile and robust AI systems.

\renewcommand{\thesection}{\Alph{section}}
\setcounter{section}{0}



\section*{Appendix}

\section{More Details of \methodname Method}
\label{sec:supp_method}
\subsection{ModEmbed for Modality Encoder}
\label{sec:supp_modemb}
As describe in~\cref{sec:arch}, we adopt a specific tokenization scheme to transform raw input signals into token embeddings for each modality. In this section, we introduce the modality embedding modules for 3D point cloud, depth, audio, tactile and EEG in our work.
\par\head{3D point cloud.} For 3D point cloud embedding, we utilize the approach introduced in~\cite{yu2022pointbert}. We initially sample $g$ center points from the input point cloud $p$ using farthest point sampling (FPS). Subsequently, we utilize the k-nearest neighbors (kNN) algorithm to select $k$ nearest neighbor points for each center point, forming $g$ local patches $\left\{ p_i \right\}_{i=1}^{g}$. To extract the structural patterns and spatial coordinates of these local patches, we normalize them by subtracting their center coordinates. Further, we employ a mini-PointNet~\cite{qi2017pointnet} to project these sub-clouds into point embeddings. Additionally, we incorporate learnable positional embeddings on top of these embeddings, serving as inputs to standard Transformers or Lens models.

\par\head{Audio.} For audio embedding, following~\cite{gong21bast}, we firstly convert the input audio waveform into a sequence of log Mel filterbank (fbank) features, forming a spectrogram with time and frequency dimensions. This spectrogram is then partitioned into a sequence of $P \times P$ patches with a stride of $S$ in both time and frequency dimensions. Each $P \times P$ patch is flattened and projected into a 1D embedding of size $d$ using a linear projection layer. Subsequently, we introduce learnable positional embeddings to capture the spatial structure of the spectrogram. These embeddings are utilized as inputs for subsequent processing by the model.

\par\head{Depth.} For depth embedding, we firstly follow~\cite{girdhar2022omnivore,girdhar2023imagebind} to convert depth maps into disparity for scale normalization. We then utilize patch embedding similar to the mechanism in \vit. This involves partitioning the disparity into $P \times P$ patches with a stride of $S$ ($S = P$) to handle the single-channel input. Each $P \times P$ patch undergoes flattening and projection into a 1D embedding of size $d$ using a linear projection layer. To capture positional information, we incorporate learnable positional embeddings. These embeddings serve as inputs for the subsequent module.

\par\head{Tactile.} For tactile embedding, since we use RGB data from GelSight~\cite{johnson2009gelsight}, we apply the same patch embedding as in \vit. Specifically, we partition the RGB input into $P \times P$ patches with a stride of $S$ ($S = P$). Each $P \times P$ patch undergoes flattening and projection into a 1D embedding of size $d$ using a linear projection layer. 
We integrate learnable positional embeddings for position information. These embeddings are forwarded as inputs for the subsequent module.

\par\head{EEG.} For EEG embedding, we use the $C$ channel EEG with $T$ timestamps. We then group every $t$ time steps into a token and transformed it into a $d$-dimensional embedding. We further add positional embeddings on top and use the yielded embeddings as inputs for the subsequent module. 

\subsection{More Details for Lens}
\par\head{Reducing computational comprexity with Iter-CS-Attn.} 
As is shown in~\cref{fig:lens_variant} in the main paper, 
the cross attention mechanism generates an output with equal length of the latent query input. 
In practice, we typically configure the Lens with less parameters (fewer attention layers) compared to the \ptvit component. Consequently, the majority of computational overhead is incurred during the forward pass of the \vit layers.
Consider the input latent query length $\ve{n}$ and the modality embedding length $\ve{m}$. For modalities with lengthy input ($\ve{m}>\ve{n}$), utilizing the Iter-CS-Attn Lens reduces the computational cost of \ptvit to $\mathcal{O}(\ve{n}^2)$, compared to encoding embeddings of the same length as the input, which has a complexity of $\mathcal{O}(\ve{m}^2)$. This strategy significantly lowers the computational overhead for processing lengthy inputs.

\subsection{Utilizing Pretrained ViT Layers}
The core of enhancing omni-modal representation with \methodname is to leverage the rich knowledge encoded in the \vit that is pretrained on large-scale data. 
To integrate \ptvit into the modality encoder, we apply the last $l$ out of the total $L$ transformer layers while maintaining a relatively high ratio $\frac{l}{L}$. 
This strategy draws inspiration from recent research exploring \vit interpretation~\cite{raghu2021vit_interprete_1,ghiasi2022vit_interprete_2}. These studies revealed that \vit captures higher-level semantic concepts in its deeper layers while encoding general edges and textures in the shallower ones.
Building upon these insights, we posit that the shared high-level knowledge among different modalities is mostly preserved in the deeper layers of the \vit architecture. Consequently, we propose the utilization of a set of \ptvit layers within the modality encoder in our pipeline. 
Notably, when $\frac{l}{L} < 1$, we either discard the initial $L - l$ transformer layers or integrate them for S-Attn type Lens learning  if applicable.

\section{More Experimental Details and Results}
\label{sec:supp_exp}
\subsection{Datasets and Metrics}
\par\head{\threeDIcon{}ULIP-ShapeNet Triplets}~\cite{xue2023ulip}.
The ULIP-ShapeNet Triplets training data for 3D point cloud is derived from ShapeNet55~\cite{chang2015shapenet} by Xue \etal~\cite{xue2023ulip}. All the 3D point clouds are generated from CAD models. Anchor images are synthesized using virtual cameras positioned around each object, and texts are obtained by filling metadata into a predefined prompt template. This dataset comprises approximately 52.5k 3D point cloud instances.

\par\head{\threeDIcon{}ULIP2-Objaberse Triplets}~\cite{xue2023ulip2}.
The ULIP2-Objaverse Triplets training data for 3D point cloud is developed by Xue \etal~\cite{xue2023ulip2}, utilizing the recently released Objaverse~\cite{deitke2023objaverse}. For each 3D object, 12 rendered images are obtained, spaced equally by 360/12 degrees. Each rendered image has 10 detailed captions generated using BLIP2-opt6.7B~\cite{blip2}. It includes around 798.8k 3D point cloud instances.

\par\head{\threeDIcon{}OpenShape Triplets}~\cite{liu2023openshape}.
The OpenShape Triplets training data for 3D point clouds encompasses four prominent public 3D datasets: ShapeNet~\cite{chang2015shapenet}, 3D-FUTURE~\cite{fu2021-3d-future}, ABO~\cite{collins2022abo} and Objaverse~\cite{deitke2023objaverse}. For each 3D object, 12 color images are rendered from preset camera poses, and thumbnail images are included as candidates if provided. OpenShape employs various strategies to obtain high-quality text descriptions, including filtering noisy metadata using GPT4~\cite{openai2023gpt4}, generating captions using BLIP~\cite{blip} and Azure cognition services, and conducting image retrieval on LAION-5B to retrieve relevant texts with paired images closely resembling the object's rendered image, leading to a wider range of text styles. This dataset comprises approximately 876k 3D point cloud instances.

\par\head{\threeDIcon{}ModelNet40}~\cite{wu2015modelnet}. 
The ModelNet40 dataset is a widely used benchmark in the field of 3D object recognition. It consists of 12,311 CAD models from 40 categories, with 9,843 training samples and 2,468 testing samples. It includes everyday objects such as chairs, tables, desks, and other household items. Each object is represented as a 3D point cloud and has been manually annotated with the object's category. The dataset is commonly used for tasks like shape classification and shape retrieval. In this work, we only use the test samples for zero-shot classification. The evaluation is performed using Top-K accuracy.

\par\head{\threeDIcon{}ScanObjectNN}~\cite{uy2019scanobjectnn}.
The ScanObjectNN dataset is a significant resource in the domain of 3D object recognition and segmentation. It encompasses a diverse array of 3D object instances acquired through a commodity RGB-D camera. This dataset exhibits a wide spectrum of household items, furniture, and common indoor objects. Each individual object instance is annotated with fine-grained semantic and instance-level labels. In total, it contains 2,902 objects distributed across 15 distinct categories.
In this work, we follow~\cite{liu2023openshape} to use the variant provided by~\cite{yu2022pointbert} for zero-shot classification, which contains 581 test shapes with 15 categories. The evaluation is performed using Top-K accuracy.

\par\head{\threeDIcon{}Objaverse-LVIS}~\cite{deitke2023objaverse}. This dataset is an annotated subset of Objaverse~\cite{deitke2023objaverse} and consists of 46,832 shapes from 1,156 LVIS~\cite{gupta2019lvis} categories. With a larger base of classes compared to other benchmarks, Objaverse-LVIS presents a challenging long-tailed distribution, making it a better reflection of the model's performance in open-world scenarios. In this work, we follow~\cite{liu2023openshape} to use this dataset for zero-shot classification, and the evaluation is performed using Top-K accuracy.

\par\head{\depthIcon{}SUN-RGBD}~\cite{song2015sun-rgbd}. We utilize paired RGB and depth maps along with associated class labels from the SUN-RGBD dataset. For training \methodname, we employ the train set comprising approximately 5k samples.
To evaluate classification performance, we use the test set ({\bf SUN Depth-only}), which contains 4,660 samples. Specifically for testing, we only utilize depth as input and construct classification templates using the 19 scene classes available in the dataset. The evaluation process involves Top-K accuracy metrics.

\par\head{\depthIcon{}NYU-Depth v2}~\cite{Silberman2012nyudepthv2}.
We utilize the depth maps from NYU-Depth v2 test set ({\bf NYU-v2 Depth-only}) containing 654 samples for evaluation. We use 16 semantic classes in the dataset and follow previous work~\cite{girdhar2023imagebind} to conduct 10-class classification. Concretely, for classification, there is an ``others'' class corresponding to 7 different semantic classes -- [`computer room', `study', `playroom', `office kitchen', `reception room', `lobby', `study space']. For classification, we compute the similarity of the ``others'' class as the maximum cosine similarity among these 7 class names. We report Top-K accuracy.

\par\head{\audioIcon{}Audioset}~\cite{audioset}.
This dataset is utilized for both training and evaluation in our work. It contains 10-second videos sourced from YouTube and is annotated across 527 classes. It consists of 3 pre-defined splits -- unbalanced-train split with about 2M videos, balanced-train with about 20k videos and test split with about 18k videos. 
Due to the unavailability of some videos for download, we finally have 1.69M/18.7k/17.1k for these three splits. We use the train splits for training and the test split for evaluation. During evaluation and when textual data serves as anchor data during training, we make use of the textual class names along with templates. The evaluation metric employed is mean Average Precision (mAP).

\par\head{\audioIcon{}ESC 5-folds}~\cite{audioset}.
The ESC50 dataset is a widely used benchmark dataset in the field of environmental sound classification. It comprises a collection of 2,000 sound recordings, categorically organized into 50 classes, including animal vocalizations, natural soundscapes, and human-made sounds. Each class in the dataset contains 40 audio samples that are five seconds long. It has pre-defined 5 fold evaluation, each consisting of 400 test audio clips. In this work, we evaluate the zero-shot prediction on across the 5 folds and report the overall Top-1 accuracy.

\par\head{\audioIcon{}Clotho}~\cite{drossos2020clotho}.
The Clotho dataset is an audio collection paired with rich textual descriptions, comprising a development set of 2,893 audio clips and a test set of 1,045 audio clips. Each audio clip is associated with five descriptions. In this study, we focus on the text-to-audio retrieval task. For evaluation, we treat each of the five associated captions as an individual test query, searching within the set of audio clips. We employ recall@K as the evaluation metric, where a query is considered successful if the ground truth audio is retrieved among the top-K returned audio clips.

\par\head{\audioIcon{}AudipCaps}~\cite{audiocaps}.
This dataset comprises audio-visual clips sourced from YouTube, accompanied by textual descriptions. It features clips extracted from the Audioset dataset. In this study, we employed the dataset splits outlined in~\cite{oncescu2021audiocaps_exclude_as}, specifically excluding clips that intersected with the VGGSound dataset.   We end up with 813 clips in the test split for zero-shot evaluation. The task is text-to-audio retrieval and is evaluated by the recall@K metric.

\par\head{\audioIcon{}VGGSound}~\cite{chen2020vggsound}.
This is an audio-visual dataset sourced from YouYube. It contains more around 200k video clips of 10s long. These clips are annotated into 309 classes, covering a spectrum from human actions to sound-emitting objects and human-object interactions. 
Since some videos are no long available for downloading, we finally end up with 162k clips for train set and 15.5k for test set. In this work, the audio from the test set is utilized specifically for zero-shot classification tasks, evaluating model performance using the Top-1 accuracy metric.

\par\head{\tactileIcon{}Touch-and-go}~\cite{yang2022touch_and_go}.
The Touch-and-Go dataset comprises real-world visual and tactile data gathered by human data collectors probing objects in natural settings using tactile sensors while simultaneously recording egocentric video. It offers annotations for 20 material classes, and provide hard/soft (H/S) and rougH/Smooth (R/S) labels. 
The dataset is organized into distinct splits: train-material and train-H/S with 92k samples, test-material and test-H/S with 30k samples, train-R/S with 35k samples and test-R/S with 8k samples. 
In our work, we utilize the train-material split for training and perform classification on the test-material subset. For zero-shot classification, we employ test-H/S and test-R/S subsets. In the context of linear probing, we fine-tune the model using the corresponding train set for a particular task. Our evaluation of model performance utilizes the Top-1 accuracy metric.

\par\head{\tactileIcon{}ImageNet-EEG}~\cite{spampinato2017eeg_data}.
This dataset comprises EEG recordings obtained from six subjects while they were presented with 2,000 images across 40 categories from the ImageNet dataset~\cite{imagenet}. Each category contains 50 distinct images, resulting in a total of 12,000 128-channel EEG sequences.
Recorded using a 128-channel Brainvision EEG system, the dataset covers diverse object categories, including animals (such as dogs, cats, elephants), vehicles (including airliners, bikes, cars), and everyday objects (such as computers, chairs, mugs).
We leverage the observed image and/or its corresponding text label as anchor data. We conduct classification tasks on both the validation set (consisting of 1,998 samples) and the test set (consisting of 1,997 samples).
Our evaluation of model performance is based on the Top-1 accuracy metric.

\subsection{Data Input and Augmentation}
\par\head{Image and Video}.
 When handling modalities such as images, videos, or tactile sensor data with RGB or RGBT inputs, we adopt the standard input representation used in the vanilla \vit model. Specifically, for image input, we partition it into patches of size $P \times P$. For video input, we employ 2-frame clips similar to the approach outlined in~\cite{girdhar2023imagebind}. We construct patches of size $T \times P \times P$. Notably, $T = 2$, $P=16$ for \vitlensB, and $P=14$ for \vitlensL and \vitlensG. We inflate the visual encoder's weights to to accommodate spatiotemporal patches for video inputs. During inference, we aggregate features over multiple 2-frame clips. This adaptation enables models initially trained on image-text data to effectively handle videos.

\par\head{3D point cloud}. For 3D point cloud input, we follow previous work to uniformly sample 8,192 points~\cite{xue2023ulip,xue2023ulip2} or 10,000 points~\cite{liu2023openshape} as the input for 3D shape. 
During training, we apply standard augmentation~\cite{xue2023ulip} for the point clouds. As mentioned in~\cref{sec:supp_modemb}, we construct local patches by sampling 512 sub-clouds, each comprising 32 points. This is accomplished by employing Farthest Point Sampling (FPS) and the k-Nearest Neighbors (kNN) algorithm.

\par\head{Depth}.
For the single-view depth, we follow~\cite{girdhar2023imagebind} to use the in-filled depth and convert them into disparity. During training, when image is used as anchor data, we apply strong data augmentation for the anchor image, including RandAug~\cite{cubuk2020randaugment} and RandErase~\cite{zhong2020randomerase}. We used aligned spatial crop for image and depth. For embedding module, we follow the \clipvit to set $P=16$ for \vitlensB and $P=14$ for \vitlensL.

\par\head{Audio}. For audio input, we process each raw audio waveform by sampling it at 16kHZ, followed by extracting a log mel spectrogram with 128 frequency bins using a 25ms Hamming window with the a hop length of 10ms. Consequently, for an audio duration of $t$ seconds, our input dimensionality becomes $128 \times 100t$.
During training, we randomly sample a 5-second clip for audio input, and apply spectrogram masking~\cite{park2019specaugment} with max time mask length of 48 frames and max frequency mask length of 12 bins.
When image is used as anchor data, we randomly sample 1 frame from the corresponding clip and apply RandAug~\cite{cubuk2020randaugment} for the sampled frame. We also apply Mixup~\cite{zhang2017mixup} during training for both audio and its anchor data, with a mixup ratio of 0.5.
For embedding module, we set $P=16$ for \vitlensB and $P=14$ for \vitlensL, and $S = 10$.
At inference time, we uniformly sample multiple clips to cover the full length of the input sample and aggregate the features extracted from these clips.

\par\head{Tactile}.
For data from tactile sensors, we treat it similarly to RGB images. During training, we introduce random flips along the horizontal and vertical directions to augment the tactile input. Additionally, random rotations are applied to further augment the input data.
When image is used as anchor data for training, we apply RandAug~\cite{cubuk2020randaugment} to augment the image.
For embedding module, we follow the \clipvit to set $P=16$ for \vitlensB and $P=14$ for \vitlensL.

\par\head{EEG}. For EEG input data, we follow~\cite{bai2023dreamdiffusion} to use the 128-channel EEG sequences. These EEG signals are filtered within the frequency range of 5-95Hz and truncated into a common length of 512.
For embedding module,we utilize Conv1D, configuring the kernel size to 1 and the stride to 1.

\begin{table*}[t]
\centering
\begin{minipage}{\textwidth}
\centering
\resizebox{.95\textwidth}{!}{%
\setlength{\tabcolsep}{5pt}
\begin{tabular}{cccccc}
\multicolumn{1}{c|}{}                                                                 & \begin{tabular}[c]{@{}c@{}}\threeDIcon{}\textit{3D} \\ \textit{Point Cloud}\end{tabular} & \depthIcon{}\textit{Depth}                    & \audioIcon{}\textit{Audio}                & \tactileIcon{}\textit{Tactile}                 & \eegIcon{}\textit{EEG}                  \\ \Xhline{2\arrayrulewidth}
\multicolumn{1}{r|}{ModEmbed $\blacktriangleright$}                                                         & Mini PointNet                                             & PatchEmbed               & PatchEmbed           & PatchEmbed              & Conv1D               \\
\rowcolor[HTML]{ECF4FF} 
\multicolumn{1}{c|}{\cellcolor[HTML]{ECF4FF}}                                         & Iter-CS-Attn                                              & S-Attn                   & Iter-CS-Attn         & S-Attn                  & Iter-CS-Attn         \\
\rowcolor[HTML]{ECF4FF} 
\multicolumn{1}{c|}{\cellcolor[HTML]{ECF4FF}}                                         & $N=4, M=1$                                                  & $N=4$                      & $N=2, M=3$             & $N=4$                     & $N=1, M=1$             \\
\rowcolor[HTML]{ECF4FF} 
\multicolumn{1}{r|}{\multirow{-3}{*}{\cellcolor[HTML]{ECF4FF}Lens Config $\blacktriangleright$}}            & \cmark tie weights                                               & \clipvit \texttt{Block.1-4} Init & -                    & \clipvit \texttt{Block.1-4} Init & -                    \\
\multicolumn{1}{r|}{\begin{tabular}[r]{@{}c@{}}Pretrained ViT \\ Config\end{tabular} $\blacktriangleright$} & \begin{tabular}[c]{@{}c@{}}\clipvit \\ \texttt{Block.1-12}\end{tabular} & \begin{tabular}[c]{@{}c@{}}\clipvit \\ \texttt{Block.5-12}\end{tabular} & \begin{tabular}[c]{@{}c@{}}\clipvit \\ \texttt{Block.1-12}\end{tabular} & \begin{tabular}[c]{@{}c@{}}\clipvit \\ \texttt{Block.5-12}\end{tabular} & \begin{tabular}[c]{@{}c@{}}\clipvit \\ \texttt{Block.1-12}\end{tabular} \\ \bottomrule
\multicolumn{1}{l}{}                                                                  & \multicolumn{1}{l}{}                                      & \multicolumn{1}{l}{}     & \multicolumn{1}{l}{} & \multicolumn{1}{l}{}    & \multicolumn{1}{l}{} \\
\multicolumn{1}{l}{\vitlensB}                                                          &                                                           &                          &                      &                         &                      \\ \hline
\multicolumn{1}{c|}{\# Trainable Param.}                                              & 34.1M                                                     & 28.7M                    & 72.1M                & 29.1M                   & 17.4M                \\
\multicolumn{1}{c|}{\# Total Param.}                                                  & 119.7M                                                    & 85.9M                    & 157.7M               & 86.2M                   & 103.0M               \\
\multicolumn{1}{c|}{Flops}                                                            & 75.4G                                                     & 36.5G                    & 64.7G                & 36.6G                   & 41.1G                \\
\multicolumn{6}{l}{\vitlensL}                                                                                                                                                                                                                         \\ \hline
\multicolumn{1}{c|}{\# Trainable Param.}                                              & 60.0M                                                     & 50.9M                    & 127.6M               & 51.3M                   & 30.6M                \\
\multicolumn{1}{c|}{\# Total Param.}                                                  & 363.4M                                                    & 303.8M                   & 431.0M               & 304.0M                  & 333.9M               \\
\multicolumn{1}{c|}{Flops}                                                            & 236.7G                                                    & 168.6G                   & 233.6G               & 168.8G                  & 183.7G              
\end{tabular}
}
\caption{
{\bf Model Configuration for \methodname.} We show the model configurations for the modality encoder across 3D point cloud, depth, audio, tactile, and EEG, for both \vitlensB and \vitlensL architectures. 
For modality embedding module, we list the name of architecture. 
For modality Lens configuration, we specify the adopted Lens type. For S-Attn type, $N$ denotes the number of self-attention layers, accompanied by details on weight initialization. For Iter-CS-Attn type,  $N$ represents the number of basis blocks and $M$ denotes the number of self-attention layers within each basis block. The term ``tie weights'' means parameter sharing among blocks $\geq 2$~\cite{jaegle2021perceiver}.
For the \ptvit configuration, we showcase the set of frozen transformer layers used in the modality encoder.
With the listed configurations, we show the number of trainable parameters, the number of total parameters and Flops for each modality encoder.
}
\label{tab:supp_model_config}
\vspace{1em}
\end{minipage}
\hfill
\begin{minipage}{\textwidth}
\centering
\resizebox{\textwidth}{!}{
\setlength{\tabcolsep}{2pt}
\begin{tabular}{l|ccccc}
 & \begin{tabular}[c]{@{}c@{}}\threeDIcon{}\textit{3D}\\ \textit{Point Cloud}\end{tabular}  & \depthIcon{}\textit{Depth}  & \audioIcon{}\textit{Audio}  & \tactileIcon{}\textit{Tactile} & \eegIcon{}\textit{EEG} \\ \Xhline{3\arrayrulewidth}
Optimizer          & \multicolumn{5}{c}{AdamW}   \\
Optimizer momentum & \multicolumn{5}{c}{$\beta_1 = 0.9, \beta_2 = 0.98$}  \\
Peak LR            & 5e-4/5e-4/2e-4$^\star$    & 2e-4     & 2e-4  & 2e-4                         & 2e-4     \\
Weight decay       & \multicolumn{5}{c}{0.2$^\diamond$}                                  \\
Batch size         & 512                                                   & 512                              & 2048                                                & 512                          & 512      \\
Warmup steps       & \multicolumn{5}{c}{10,000}                               \\
Sample replication & 1                                                     & 50                               & 1                                                   & 50                           & 50       \\
Total epochs       & 200/150/150$^\star$                                           & 100                              & 80                                                  & 40                           & 40       \\
\rowcolor[HTML]{ECF4FF} 
Modality augmentation     & \begin{tabular}[c]{@{}c@{}} RandDropout\\  RandScale\\  RandShift\\  RandPerturb\\  RandRotate\end{tabular}$^\ast$ & \begin{tabular}[c]{@{}c@{}} RandResizeCrop(size=224)\\  RandHorizontalFlip(p=0.5)\\  RandAugment(m=9, n=2)\\  RandErasing(p=0.25)\end{tabular}                   & \begin{tabular}[c]{@{}c@{}}Frequency masking(12)\\ Time masking(48)\\ NoiseAug\\ mixup(p=0.5)\end{tabular}          & \begin{tabular}[c]{@{}c@{}} RandResizeCrop(size=224)\\  RandHorizontalFlip(p=0.5)\\  RandVerticalFlip(p=0.5)\\  RandRotation(degrees=(0,360))\end{tabular} & -        \\
\rowcolor[HTML]{EFEFEF} 
Image augmentation &  RandResizeCrop(size=224)$^\ast$                             & \begin{tabular}[c]{@{}c@{}} RandResizeCrop(size=224)\\  RandHorizontalFlip(p=0.5)\\  RandAugment(m=9,n=2)\\ ColorJitter(0.4)\\  RandErasing(p=0.25)\end{tabular} & \begin{tabular}[c]{@{}c@{}}RandShortSideScale(min=256, max=340)\\  RandCrop(size=224)\\  RandHorizontalFlip(p=0.5)\\  RandAugment(m=9, n=2, p=0.3)\\ mixup(p=0.5)\end{tabular} & \begin{tabular}[c]{@{}c@{}} RandResizeCrop(size=224)\\  RandHorizontalFlip(p=0.5)\\  RandAugment(m=9,n=2)\\ ColorJitter(0.4)\end{tabular}                     & \begin{tabular}[c]{@{}c@{}} RandResizeCrop(size=224)\\  RandHorizontalFlip(p=0.5)\\  RandAugment(m=9,n=2)\\ ColorJitter(0.4)\end{tabular}
\end{tabular}
}
\caption{{\bf Training hyper-parameters for each modality.}
$^\star$ Separate hyper-parameters are reported for 3D training with different datasets: ULIP-ShapeNet, ULIP2-Objaverse, and OpenShape Triplets.
$^\ast$ Augmentations listed for 3D training are applied to ULIP-ShapeNet and ULIP2-Objaverse, while released features are used for training on OpenShape Triplets.
$^\diamond$ Weight decay excludes parameters for BatchNorm, LayerNorm, bias terms, and logit scale.
}\label{tab:supp_train_setup}
\end{minipage}
\vspace{-1em}
\end{table*}

\subsection{Model Configuration}
In this section, we provide the configurations for encoders of different modalities in \methodname. 
Details are specified in~\cref{tab:supp_model_config}.

\subsection{Training Setup}
In~\cref{tab:supp_train_setup}, we list the hyper-parameters used in training for each modality. Our experiments were done on 32GB V100 GPU clusters.

\subsection{Additional Results on Understanding Tasks}
\par\head{Additional results on audio tasks.} 
We merged Audioset~\cite{audioset} and VGGSound~\cite{chen2020vggsound}, resulting in a combined training dataset with 1.86M samples. Results in~\cref{tab:supp-additional-audio} show improved performance across all benchmarks due to the inclusion of additional training data.
\begin{table}[h]
\centering
\resizebox{1.\linewidth}{!}{%
        \begin{tabular}{c|c|c|c|cc|cc}
        \multirow{2}{*}{Training data} & AS-A & VGGS   & ESC   & \multicolumn{2}{c|}{Clotho } & \multicolumn{2}{c}{AudioCaps } \\
                                 & mAP      & Top1      & Top1 & R@1          & R@10         & R@1           & R@10          \\ \Xhline{3\arrayrulewidth}

        \multicolumn{1}{l|}{AS}  &   26.7     &   31.7      &   75.9 &   8.1          &   31.2         &   14.4          &   54.9 \\
        \multicolumn{1}{l|}{AS + VGGS}  &   27.2     &   51.7      &   80.9 &   7.9          &   31.5         &   14.9          &   55.2 
        
        \end{tabular}
}
\caption{{\bf Training for audio with more data.} We Audioset(AS) and Audioset+VGGSound(AS+VGGS) to train \vitlensL and report metrics following~\cref{tab:audio_cls}.}\label{tab:supp-additional-audio}
\end{table}

\subsection{More Details and Results for \Boldmethodname MFMs}
\par\head{Architectural details for \methodname MFM integration.}
Both InstructBLIP~\cite{dai2023instructblip} and SEED~\cite{ge2023seed_tokenizer,ge2023seed_llama} apply the pretrained EVA01-g14~\cite{fang2023eva} \clipvit to perceive and encode images for the subsequent LLM input. Concretely, they use the first 39 transformer layers of the 40-layer \clipvit for visual feature extraction.
Adhering to this configuration, we employ the EVA01-g14 CLIP as the foundation model and utilize its \clipvit as an integral part of the modality encoder for the training of multimodal alignment. We tune the parameters of ModEmbed and Lens.
During inference, we directly plug the ModEmbed and Lens prior to the \ptvit, enabling the yielded MFM to handle inputs of various modality without specific instruction following.

\subsubsection{Additional Results: InstructBLIP with \methodname}
\par\head{Comparison of InstructBLIP with \methodname against other methods on 3D data instruction following.}
We train \methodname for 3D point cloud using ULIP2-Objaverse and integrate it into InstructBLIP. Beyond capturing the high-level semantics of the input data, we observed that leveraging the EVA01-g14 \clipvit within the modality encoder further enhanced the model's ability to capture local details.

Our qualitative evaluation involves a comparison with: (1) PointBERT~\cite{yu2022pointbert} aligned with EVA01-g14 CLIP, replacing the vision encoder used in InstructBLIP; and (2) CLIPCap~\cite{mokady2021clipcap} from OpenShape~\cite{liu2023openshape}.
We present a snapshot of qualitative outcomes across different models in~\cref{tab:visual_example_piano}, ~\cref{tab:visual_example_plant} and ~\cref{tab:visual_example_toilet}. 
These examples showcase several capabilities exhibited by \methodname integration without specific tuning using 3D-related instructional data. Notably, the examples demonstrate that \methodname empowers InstructBLIP to accurately describe 3D objects. For instance, the plant example in ~\cref{tab:visual_example_plant} is characterized as ``sitting in a ceramic pot'' and ``bamboo-like''. Moreover, \methodname excels in capturing local visual concepts beyond the most prominent ones. For instance, the piano example in~\cref{tab:visual_example_piano} describes the observation of a ``chair''.

For PointBERT integrated InstructBLIP, although PointBERT achieves decent performance for zero-shot classification, it fails to provide accurate information for the InstructBLIP as \methodname does. We can see that in~\cref{tab:visual_example_piano}, although it recognizes the piano, it fails to provide accurate brief and detailed description since it includes ``person'' in its description, which does not exist in the 3D input. Also, it fails to recognize the plant in~\cref{tab:visual_example_plant} and the toilet in~\cref{tab:visual_example_toilet}.

CLIPCap-OpenShape, while occasionally displaying some relevant entities in captions (``vase'' in ~\cref{tab:visual_example_plant} and ``toilet'' in ~\cref{tab:visual_example_toilet}), often generates hallucinations and inaccurate captions.

The overall results demonstrate that \methodname excels not only at classifying the salient object of the 3D input, but also capturing the visual details. 
This merit is surprising: despite the fact that we only explicitly use the \texttt{[CLS]} for alignment, the integrated model exhibits the ability to capturing local information.
This ability might stem from \methodname potentially inheriting information captured by other tokens, which could carry local details to the input of InstructBLIP. This capability indicates that the model might leverage the collective knowledge present in various tokens, not limited to the \texttt{[CLS]}, contributing to its robustness in encoding rich visual information.

\begin{table*}[bp]
\begin{minipage}{0.99\textwidth}
\centering  
\scalebox{0.88}{
\begin{tabular}{l p{10.5cm} }
\toprule
 \multicolumn{2}{l}{\bf Visual input example, Piano:}  \\
\midrule
&  \includegraphics[height=3.5cm]{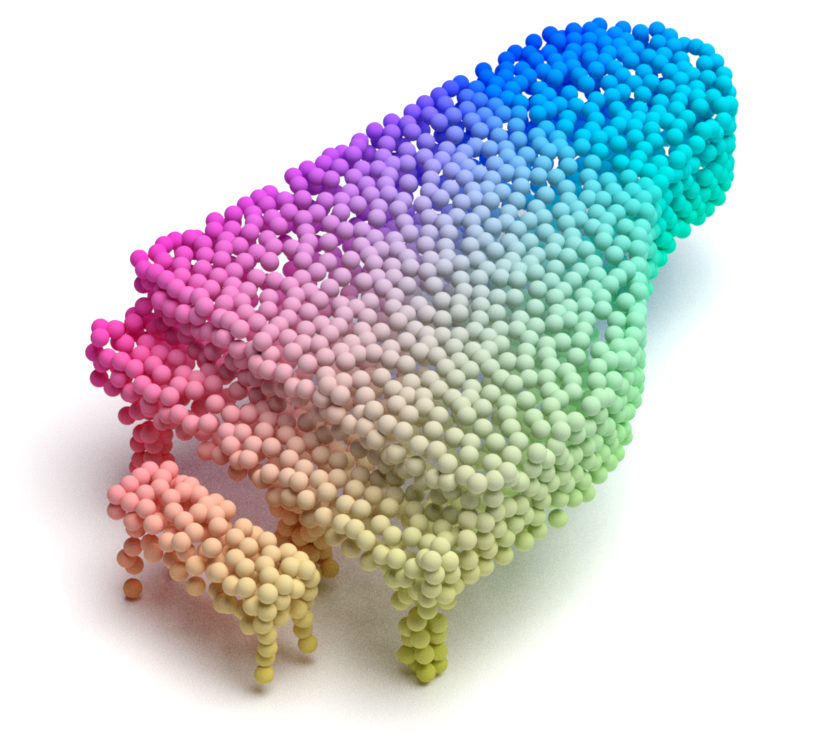} \\

User & What is this object? \\
InstructBLIP w/ \methodname & Piano.\\
User & What is the function of this? \\
InstructBLIP w/ \methodname & Pianos are used to play music.\\
User & Describe it in detail. \\
InstructBLIP w/ \methodname & The object in the image is an upright piano, which is a type of keyboard instrument known for its distinct sound and versatility. The piano is placed on a wooden surface, possibly a floor or a stage, and it appears to be \textcolor{blue}{quite large in size}. Additionally, there is \textcolor{blue}{a chair} placed next to the piano, suggesting that it may be used for seating during performances or lessons. The piano is positioned in the center of the image, making it the focal point of the scene. Overall, the combination of the piano and the chair creates an inviting atmosphere for music enthusiasts or those interested in learning how to play the piano.
\\
\midrule
User & What is this object? \\
InstructBLIP w/ PointBERT &  Advertisement. \\
User & Describe this object briefly. \\
InstructBLIP w/ PointBERT & A man sitting at a desk with a piano in the background. \\
User & Describe this object in detail. \\
InstructBLIP w/ PointBERT & This scene showcases a person sitting at a desk with a piano in the background. The person is wearing a suit and tie, indicating that they may be attending an important business meeting or event. The scene is set in a professional environment, emphasizing the importance of proper attire and etiquette. Additionally, there is a laptop on the desk, suggesting that the person may be working on various tasks related to their job or responsibilities. \\
\midrule
CLIPCap-Openshape~\cite{mokady2021clipcap,liu2023openshape} & A white house with a microphone in the middle of a dark room. \\
\bottomrule
\end{tabular}
}
\captionof{table}{Example for \methodname enabling InstructBLIP to understand 3D inputs.}  
\label{tab:visual_example_piano} 
\end{minipage}
\end{table*}
\begin{table*}[h]
\begin{minipage}{0.99\textwidth}
\centering  
\scalebox{0.88}{
\begin{tabular}{l p{10.5cm} }
\toprule
 \multicolumn{2}{l}{\bf Visual input example, Plant:}  \\
\midrule
&  \includegraphics[height=3.5cm]{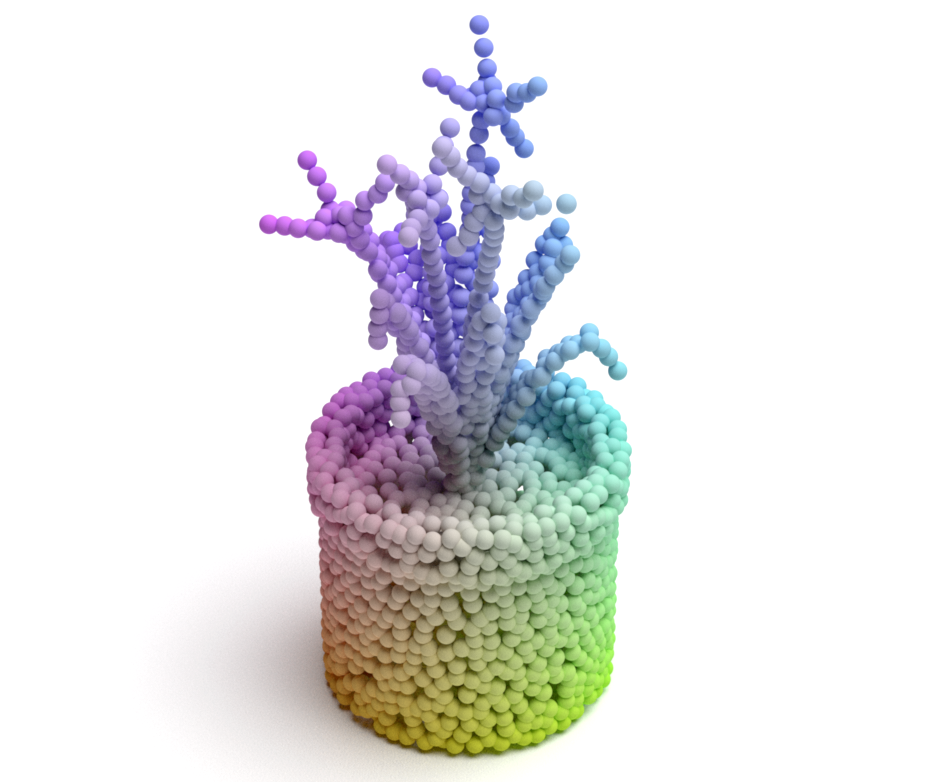} \\
User & What is this object? \\
InstructBLIP w/ \methodname & Potted plant.
\\
User & Describe it in detail. \\
InstructBLIP w/ \methodname & In the image, there is a potted plant \textcolor{blue}{sitting in a ceramic pot} on a table. The plant appears to be a \textcolor{blue}{bamboo-like} species, and it is positioned in such a way that its leaves appear to be growing out of the top of the pot. The plant's placement on the table suggests that it is meant to be a decorative element, adding a touch of greenery to the space. The ceramic pot has a diameter of around 10 centimeters, making it a suitable size for holding a small plant like this one. Overall, the image showcases a simple yet elegant arrangement of a potted plant in a ceramic pot on a table.
\\
\midrule
User & What is this object? \\
InstructBLIP w/ PointBERT &  1000px by 1000px \\
User & Describe this object briefly. \\
InstructBLIP w/ PointBERT & A man in a suit and tie is standing in front of a desk. \\
\midrule
CLIPCap-Openshape~\cite{mokady2021clipcap,liu2023openshape} & Flowers in a vase, black and white. \\
\bottomrule
\end{tabular}
} 
\vspace{-0.5em}
\captionof{table}{Example for \methodname enabling InstructBLIP to understand 3D inputs.}  
\label{tab:visual_example_plant}
\vspace{1.5em}
\end{minipage}
\hfill
\begin{minipage}{0.99\textwidth}
\centering  
\scalebox{0.88}{
\begin{tabular}{l p{10.5cm} }
\toprule
 \multicolumn{2}{l}{\bf Visual input example, Toilet:}  \\
\midrule
&  \includegraphics[height=3.5cm]{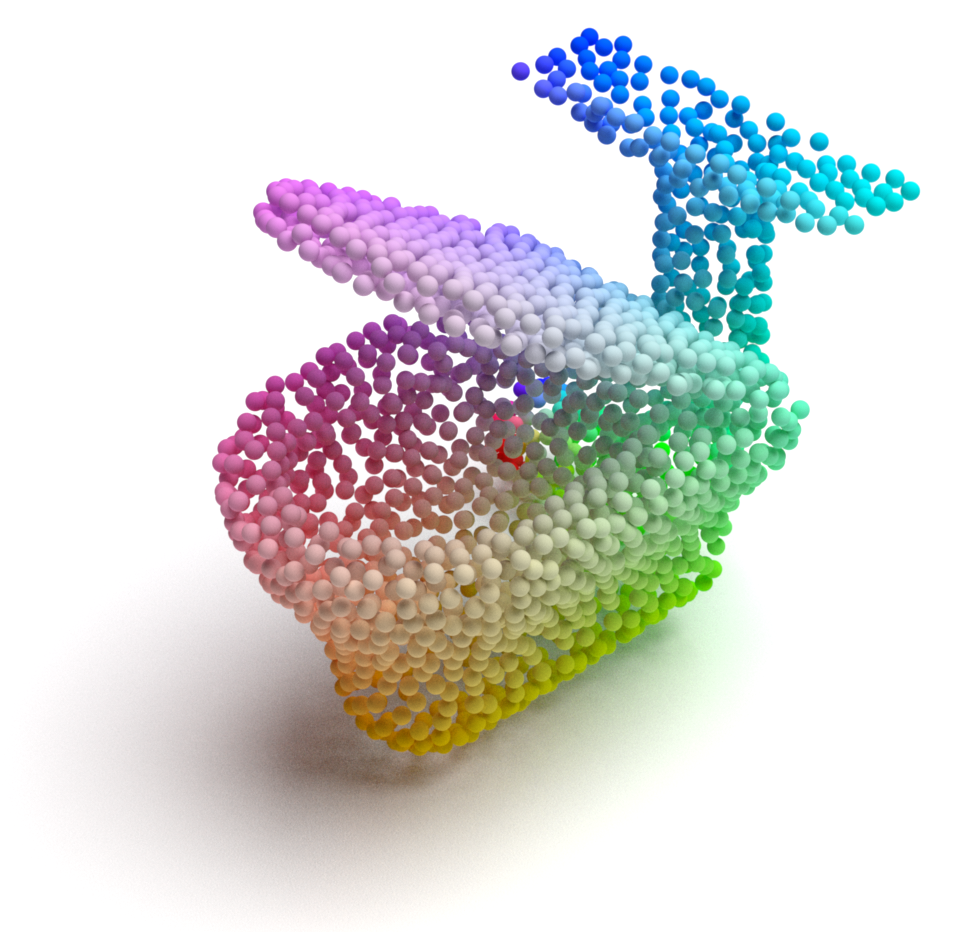} \\

User & What is in the scene? \\
InstructBLIP w/ \methodname & Toilet.\\
User & Describe it in detail. \\
InstructBLIP w/ \methodname & The image features a close-up view of a toilet with its lid open, revealing the inside of the toilet bowl. The toilet appears to be made of porcelain or ceramic material, and it has a seat attached to it. The toilet is positioned on a white background, providing a clear and detailed view of the toilet's design and construction. In addition to the toilet, there is another object in the image, which could be a piece of furniture or a decorative item, but it is not as prominent as the toilet. Overall, the image showcases a functional and well-designed toilet, emphasizing the importance of cleanliness and hygiene in daily life.
\\
\midrule
User & What is in the scene? \\
InstructBLIP w/ PointBERT &  Lamp. \\
User & Describe this object briefly. \\
InstructBLIP w/ PointBERT & An image of a desk with a lamp on it. \\
\midrule
CLIPCap-Openshape~\cite{mokady2021clipcap,liu2023openshape} & The moment a man's hand reaches out to touch a toilet bowl. \\
\bottomrule
\end{tabular}
}
\vspace{-0.5em}
\captionof{table}{Example for \methodname enabling InstructBLIP to understand 3D inputs.}  
\label{tab:visual_example_toilet} 
\end{minipage}
\end{table*}

\par\head{InstructBLIP with \methodname for input of multiple modalities.} 
We demonstrate that the versatile omni-modal \methodname encoder, coupled with an array of specialized Lenses, functions as a sensor adept at concurrently perceiving and understanding multiple modalities.  
To achieve this, we concatenate the outputs from diverse modality Lenses prior to inputting them into the \vit transformer. Subsequently, the encoded embeddings from this concatenation are forwarded to the LLM within InstructBLIP for text generation.

Qualitative results\footnote{Photos credited to \url{https://www.pexels.com/}.} are showcased in ~\cref{tab:example_mm2_insblip} for dual-modality input and in ~\cref{tab:example_mm3_insblip} for tri-modal input. The outputs produced by InstructBLIP with \methodname underscore its remarkable ability to concurrently interpret multiple modalities, akin to perceiving an image. 
Notably, as evident in the qualitative results, the incorporation of \methodname enhances the resulting MFM's capacity to digest multi-modal inputs, discover unconventional co-occurrence of concepts from different modalities, and craft stories based on the aggregated information from multiple modalities without specific instruction tuning.

\begin{table*}[h!]\centering
\begin{minipage}{0.9\textwidth}\vspace{0mm}
\centering
\begin{tcolorbox}[colback=white,colframe=blue!75!black,title={\bf InstructBLIP w/ \methodname, Two Multimodal Inputs.}]
    \centering
      \footnotesize
\begin{tabular}{p{0.9\textwidth} }
 \textcolor{blue}{ {\bf Example 1: 3D Point Cloud + Image} } \\
 \multicolumn{1}{c}{\includegraphics[height=2.0cm]{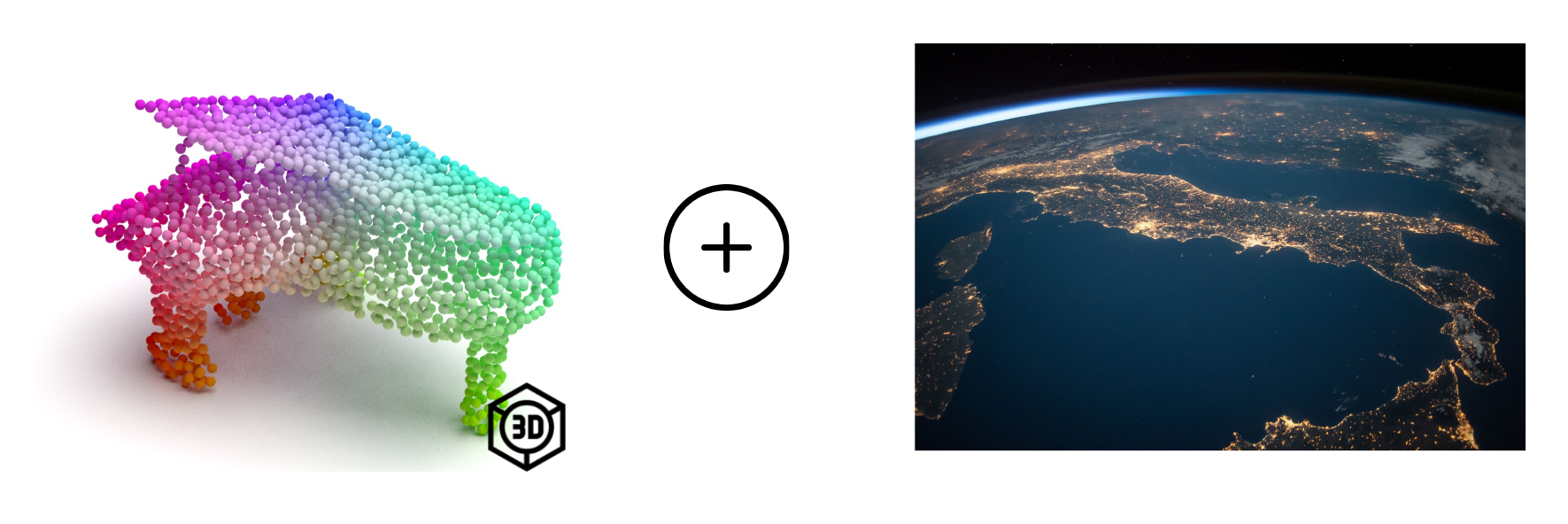}} \\
{\bf Prompt 1:} Describe based on what you see.  \\
{\bf Response 1:} An image of a piano with the earth in the background.  \\ 
{\bf Prompt 2:} What is unusual about this image?  \\
{\bf Response 2:} The unusual aspect of this image is that it depicts a piano floating in space, with the Earth visible in the background. This is an imaginative and creative representation of a musical instrument in an unconventional setting, showcasing its versatility and adaptability to different environments. \\
{\bf Prompt 3:} Imagine the events leading up to this moment and those that follow. Create a story that links them together.  \\
{\bf Response 3:} In the image, we see a grand piano placed on top of the earth, with the moon and stars visible in the background. This setting creates a unique and captivating scene that invites the viewer to imagine the events leading up to this moment and those that follow. One possible storyline could involve a talented pianist who has always dreamed of performing on top of the earth. After years of hard work and dedication, they finally have the opportunity to make their dream come true. As they prepare for their performance, they reflect on the journey that led them to this moment and the challenges they faced along the way. During their performance, they play a beautiful melody that captures the essence of the earth's beauty and the vastness of the universe. The audience is mesmerized by the performance, and the pianist feels a sense of accomplishment and fulfillment as they conclude their performance on top of the earth.  \\ \\
\hrulefill  \\ \\
\textcolor{blue}{ {\bf Example 2: 3D Point Cloud + Image} }  \\
 \multicolumn{1}{c}{\includegraphics[height=2.0cm]{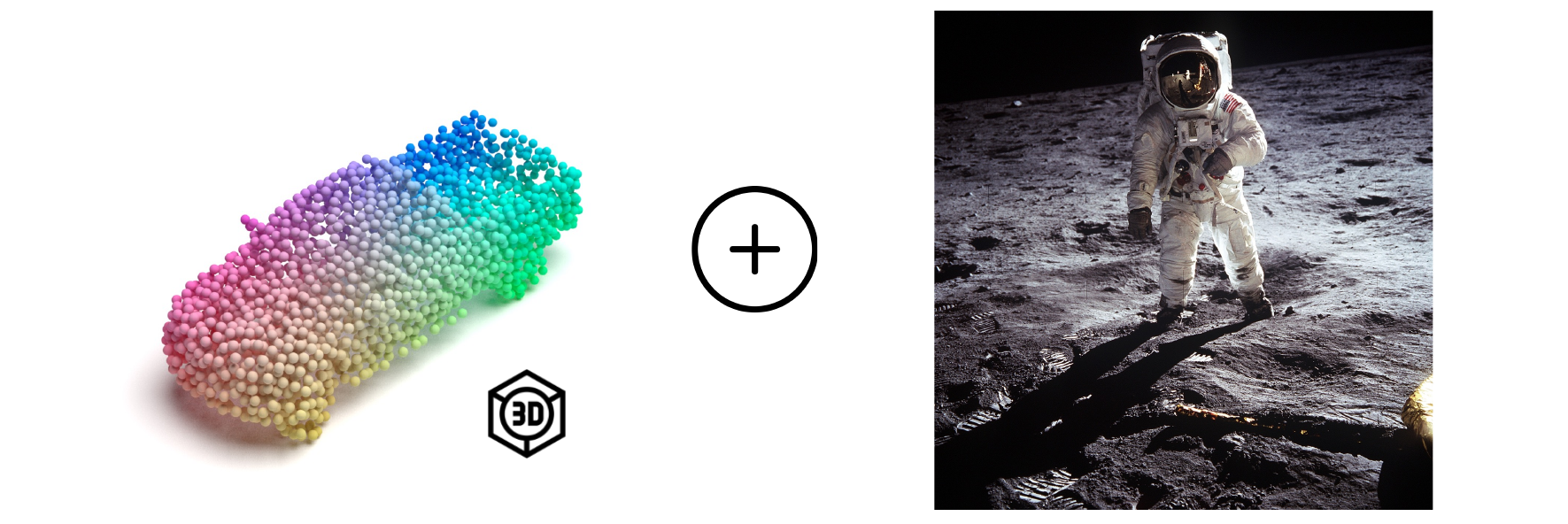}} \\
{\bf Prompt 1:} Imagine you are a narrator looking at this, describe the essence of what you see.  \\
{\bf Response 1:} An astronaut standing next to a pickup truck on the surface of the moon.  \\
{\bf Prompt 2:} What is unusual aboout this image?  \\
{\bf Response 2:} The unusual aspect of this image is that it depicts an astronaut standing on the surface of the moon with a pickup truck in the foreground. This combination of elements is not commonly associated with space exploration or the moon's surface, making it an unconventional and eye-catching image.  \\
{\bf Prompt 3:} Craft a narrative based on the visual elements in the picture. Let your imagination guide the story. \\
{\bf Response 3:} In the image, an astronaut is standing on the moon's surface with a pickup truck in the background. The astronaut is wearing a spacesuit and appears to be exploring the lunar terrain. The pickup truck is parked on the moon's surface, adding a unique touch to the scene. The combination of the astronaut and the pickup truck creates an intriguing narrative that invites the viewer to imagine what might be happening in this futuristic setting.
\\
\hrulefill  \\  
\\ 
\textcolor{blue}{ {\bf Example 3: EEG + Audio} }  \\
 \multicolumn{1}{c}{\includegraphics[height=1.8cm]{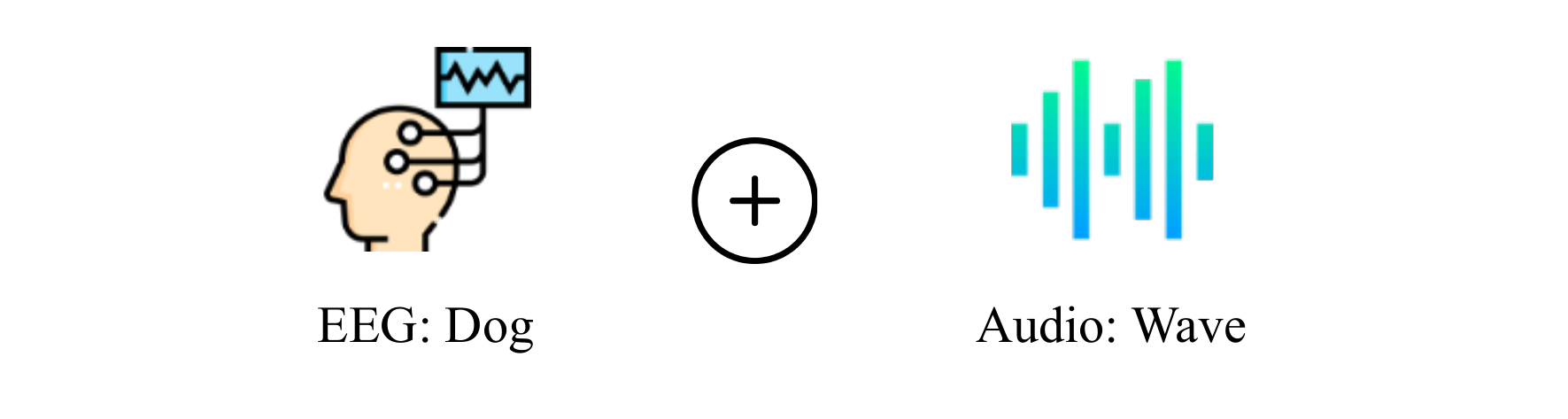}} \\
 {\bf Prompt 1:} Tell me based on what you see.  \\
{\bf Response 1:} A small dog sitting on top of a sandy beach. \\ 
{\bf Prompt 2:} Tell me more about this scene. \\
{\bf Response 2:} In the image, we see a cute little dog sitting on a sandy beach. The dog is wearing a blue collar and appears to be waiting patiently for its owner or someone else to come along. The beach is littered with small pebbles, adding to the serene atmosphere of the scene. As the sun sets in the background, the sky takes on a beautiful orange and pink hue, casting a warm and inviting glow over the entire area. The dog seems to be enjoying the peaceful surroundings, taking in the sights and sounds of the beach. It's a heartwarming image that captures the beauty of nature and the joy of spending time with a beloved pet.\\
\end{tabular}
\end{tcolorbox}
\vspace{-.5em}
\caption{Example to illustrate the Instruct-BLIP with input of two modalities.}
    \label{tab:example_mm2_insblip}
\end{minipage}
\end{table*}
\begin{table*}[h!]\centering
\begin{minipage}{0.9\textwidth}\vspace{0mm}
\centering
\begin{tcolorbox}[colback=white,colframe=blue!75!black,title={\bf InstructBLIP w/ \methodname, Three Multimodal Inputs.}]
    \centering
      \footnotesize
\begin{tabular}{p{0.9\textwidth} }
 \textcolor{blue}{ {\bf Example 1: 3D Point Cloud + Image + Image} } \\
 \multicolumn{1}{c}{\includegraphics[height=2.5cm]{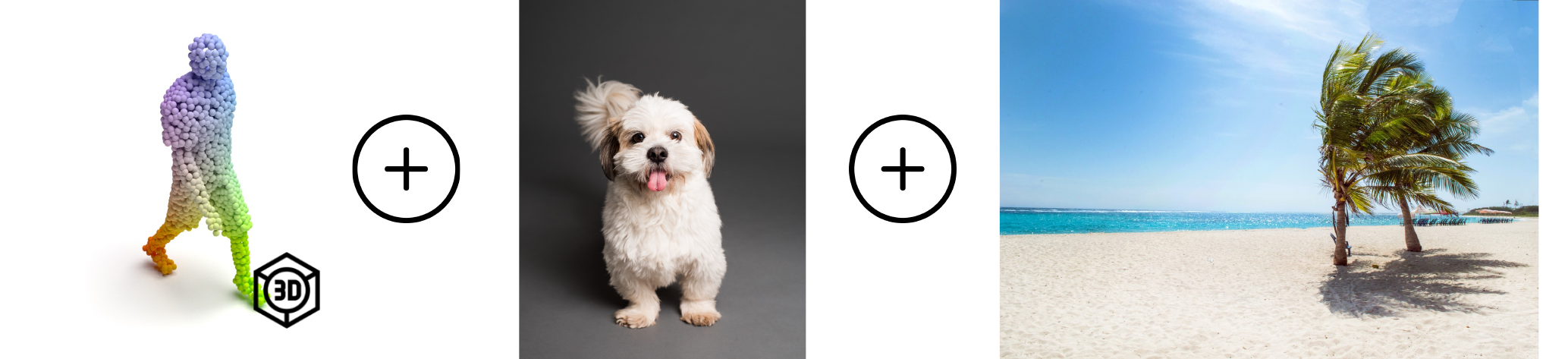}}  \\
{\bf Prompt 1:} Describe the scene based on what you see.  \\
{\bf Response 1:} A man walking a dog on a beach with palm trees in the background.  \\ 
{\bf Prompt 2:} Describe the details of what you see. \\
{\bf Response 2:} The image depicts a man and his dog walking on a sandy beach. The man is holding the dog's leash, while the dog appears to be enjoying its time on the beach. In the background, there is a palm tree visible, adding to the serene atmosphere of the scene. The man and the dog seem to be in good spirits, enjoying each other's company as they stroll along the beach.\\
{\bf Prompt 3:} Imagine the events leading up to this moment and those that follow. Create a story that links them together. \\
{\bf Response 3:} In the story, the man and his dog are walking along the beach, enjoying the sunny day. As they walk, the man notices a piece of paper lying on the sand. Curious, he picks it up and discovers that it's an advertisement for a photography contest. The man decides to enter the contest and submits a photo of himself and his dog on the beach. To his surprise, his photo is selected as one of the winners, and he is invited to attend the award ceremony. At the event, he meets other photographers and learns more about the art of photography. Through this experience, the man gains a new appreciation for photography and decides to pursue it as a hobby. He continues to take photos of his dog and other subjects, honing his skills and capturing unique moments in time. Eventually, his passion for photography leads him to become a professional photographer, and he shares his work with others, inspiring them to appreciate the beauty of the world around them. \\
\\
\hrulefill  \\  
\\ 
\textcolor{blue}{ {\bf Example 2: EEG + Tactile + 3D Point Cloud} }  \\
 \multicolumn{1}{c}{\includegraphics[height=2.0cm]{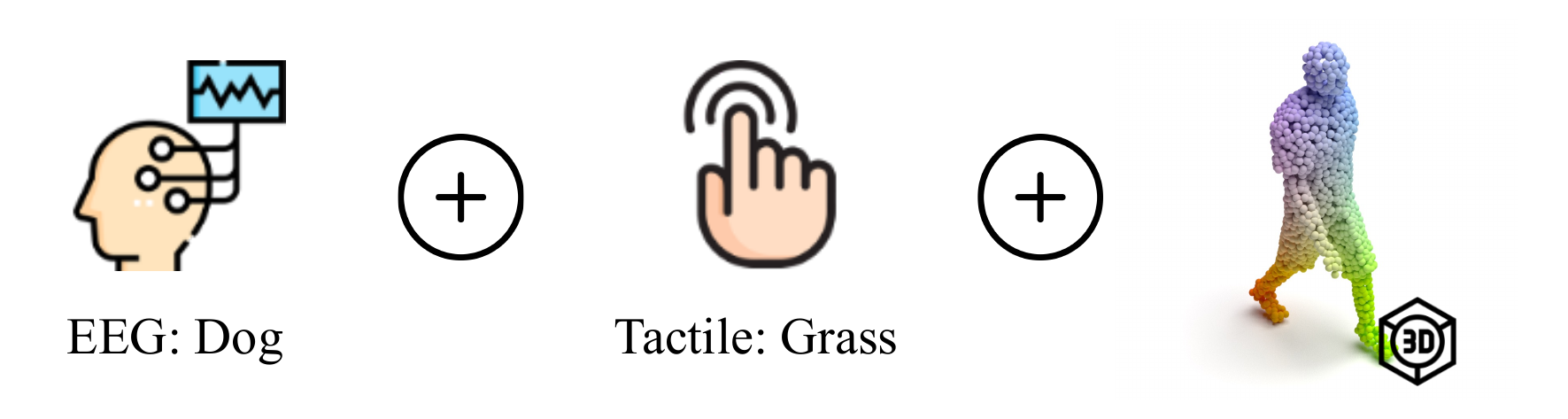}} \\
{\bf Prompt 1:} Describe what you see.  \\
{\bf Response 1:} A person and a dog walking in a grassy area.  \\
{\bf Prompt 2:} Create a short story around the given visual contents.  \\
{\bf Response 2:} This scene depicts a man walking his dog in a grassy field. The man and the dog are both visible in the scene, with the man standing on the left side of the image and the dog standing on the right side. The man is wearing a shirt and pants, while the dog is wearing a collar. The man and the dog seem to be enjoying each other's company as they walk together in the field. \\
\hrulefill  \\  
\\ 
\textcolor{blue}{ {\bf Example 3: EEG + Audio + Audio} }  \\
\multicolumn{1}{c}{\includegraphics[height=2.0cm]{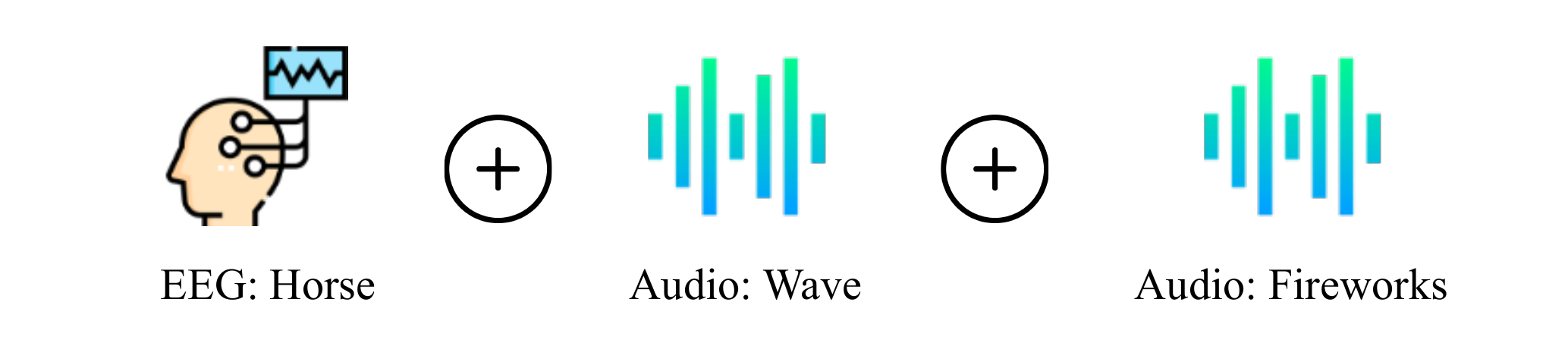}} \\
{\bf Prompt 1:} Describe what you see.  \\
{\bf Response 1:} A white horse running on the beach with fireworks in the background.  \\
{\bf Prompt 2:} Create a short story around the given visual contents.  \\
{\bf Response 2:} The image depicts a white horse standing on a beach, surrounded by water and trees. The horse appears to be in motion, possibly running or galloping across the beach. In the background, there is a firework display taking place, adding to the festive atmosphere of the scene. The combination of the horse, the beach, and the fireworks creates a unique and visually captivating image.  \\
\end{tabular}
\end{tcolorbox}
\caption{Example to illustrate the Instruct-BLIP with input of three modalities.}
    \label{tab:example_mm3_insblip}
\vspace{-2mm}
\end{minipage}
\end{table*}

\subsubsection{Additional Results: SEED with \Boldmethodname}
\begin{figure*}[h]
    \centering
    \includegraphics[height=20.8cm]{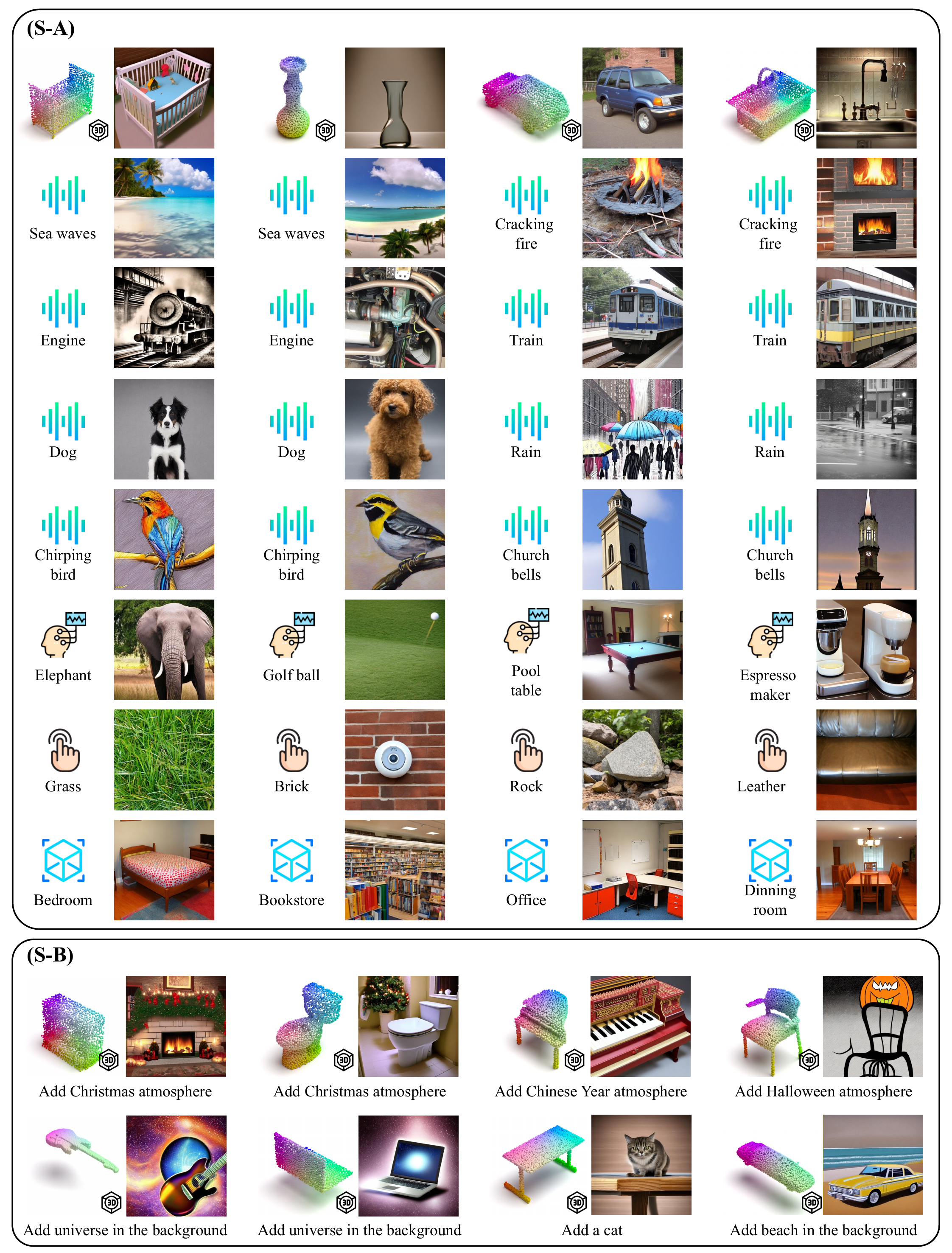}
    \vspace{-0.5em}
    \caption{
    {\bf Qualitative examples for plugging \methodname into SEED.}
    We present the input-output pairs in a local left-right pattern.
    {\bf (S-A) Any modality to image generation.} The integrated model generates an image output (right) corresponding to the provided individual input (left).
    {\bf (S-B) Compositional any modality to image generation.}  We focus on 3D point cloud cases in the examples for better visualization. The integrated model generates a corresponding image (right) when presented with the input (left) along with the conditioned text prompt.
    }
    \label{fig:supp_mm_seed}
\end{figure*}

Integrating the well-trained modality Lenses into SEED enables seamless any-modality to image generation. In addition to the primary paper, we present further qualitative results in~\cref{fig:supp_mm_seed}.

In (S-A), we showcase the capability of generating images from any modality input by encoding the modality information using Lens and the SEED tokenizer~\cite{ge2023seed_tokenizer}. We append to this with a text prompt, ``\texttt{Generate an image based on what you perceive.}'' The results demonstrate that our \methodname integrated MFM successfully generates images across diverse modalities, including 3D point clouds, audio, EEG, tactile, and depth. Notably, in the 3D point cloud examples (Row 1 in S-A), the model retains the local structure of the 3D shapes. 
Furthermore, the model exhibits the ability to generate diverse images for different inputs within the same category, exemplified by the audio examples for ``sea waves'', ``engine'', ``crackling fire'' and more. This showcases the robustness of our \methodname.

In (S-B), we show that our integration extends SEED's capability for compositional image generation to any modality. For better visualization, we show examples for 3D point clouds. We show the main instruction under each example. 
In practice, we feed any modality input into the LLM via Lens and the SEED tokenizer, supplementing it with the prompt ``\texttt{[Instruction], generate an image}'' to guide text-based generation.
The presented results highlight the model's ability to retain visual concepts accurately in the generated images. Additionally, it preserves the local structure rather than merely focusing on high-level semantics. Notably, in examples such as ``guitar'' and ``car'' the model successfully retains the shape and some local structures, showcasing its nuanced understanding beyond high-level semantic understanding.

Moreover, the model demonstrates the capability to intake inputs from various modalities and subsequently generate an image that combines all the conveyed concepts in a coherent manner. In practice, we employ the prompt ``\texttt{[input tokens A], [input tokens B], please generate an image to combine them}'' to facilitate this process. For a visual examples, please refer to~\cref{fig:vitlens_mfm}-(E) in the main paper.

\clearpage

\subsection{Applications}
The versatility of \methodname in binding diverse modalities into a unified space unlocks a multitude of applications, including cross-modal retrieval and semantic search.
This section demonstrates the application of \methodname in the domain of any-modality to 3D scene understanding, leveraging the capabilities of the recent OpenScene framework~\cite{Peng2023OpenScene}. OpenScene aligns 3D point features within the CLIP embedding space, enabling text-based and image-based searches within a 3D scene. Building upon OpenScene, \methodname extends this understanding of 3D scenes to encompass more modalities.

\begin{figure}[h]
    \centering
    \includegraphics[width=\linewidth]{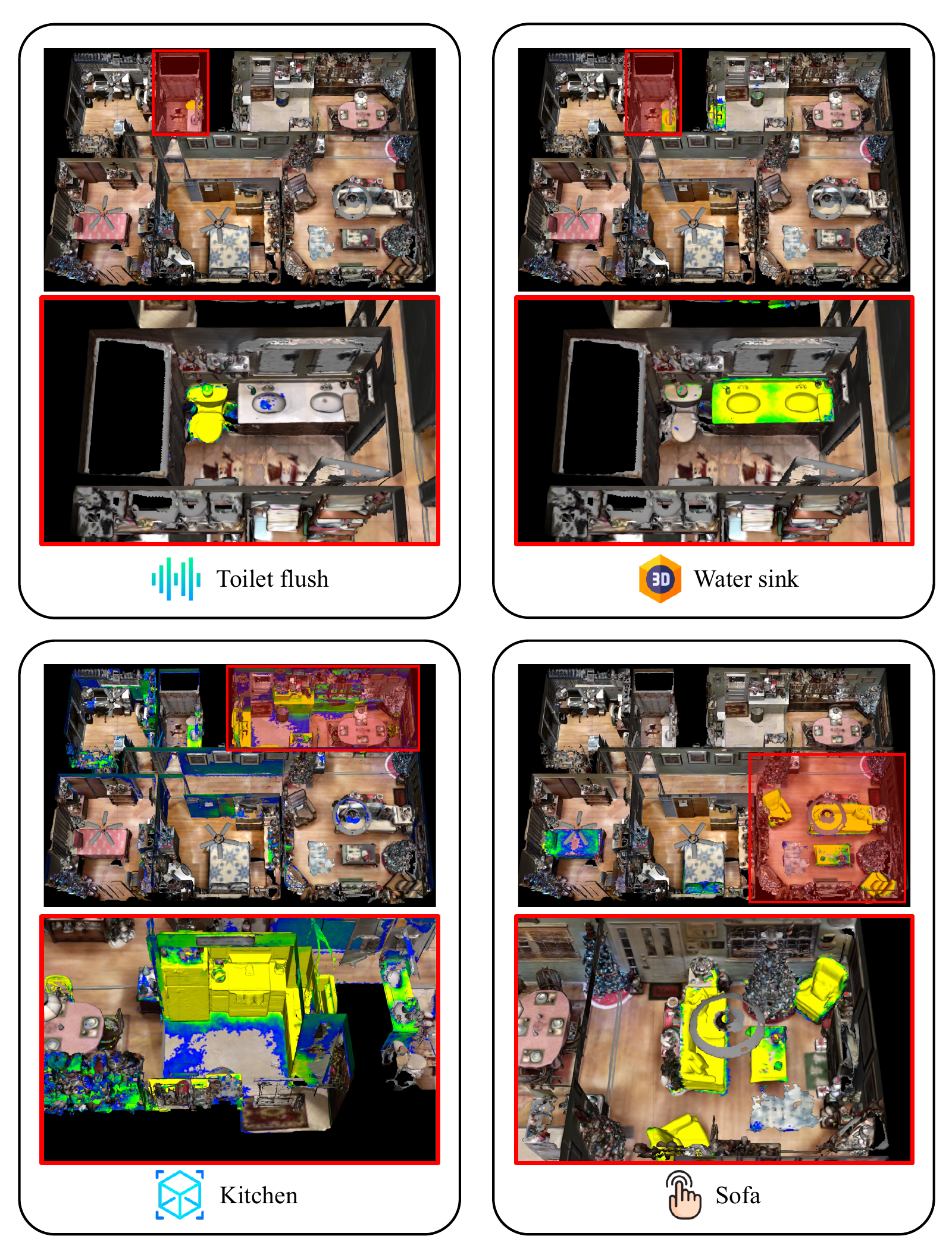}
    \caption{
    {\bf Application for any-modality to 3D scene understanding.} 
     This application facilitates scene exploration by accepting inputs from diverse modalities and subsequently highlighting relevant areas within the scene. In the visualization, the color gradient represents the relevance level within the scene (yellow is the highest, green is moderate, blue is low, and uncolored is lowest).}
    \label{fig:application_3d_scene}
\end{figure}

The qualitative results in~\cref{fig:application_3d_scene} demonstrate this application's ability to utilize inputs from multiple modalities to identify relevant areas within the scene. It effectively highlights objects like the toilet flush based on toilet audio, the sink area using 3D point cloud data of a water sink, the kitchen area from the depth map, and the presence of sofas inferred from tactile input indicating a leather sofa.

\subsection{Additional Ablation Studies}
This section presents additional ablation experiment findings regarding \methodname training.
\subsubsection{Anchor Data for Alignment}
We study the effect of using different anchor data for multimodal alignment during training. We employ \vitlensB in experiments. We train for 3D point cloud on ULIP-ShapeNet and follow the main settings for other modalities. The results are shown in~\cref{tab:supp_anchor_data}. 
Our observations reveal that employing both image and text as anchor data yields superior performance for tasks involving 3D point clouds, depth, and audio. In contrast, utilizing only image or text alone results in comparatively lower accuracy.
For tactile and EEG tasks, aligning with text produces the best results. Our speculation is that in the case of tactile data, the aligned images depict close-up views of objects, differing from those used in CLIP training. Consequently, the CLIP image encoder might not offer the optimal alignment space. 
As for EEG, due to the very limited scale of data, employing text-only alignment seems to be the most effective approach. 
\begin{table}[h]
\centering
\resizebox{0.99\linewidth}{!}{%
\setlength{\tabcolsep}{2pt}
\begin{tabular}{c|ccccc}
 Anchor data $\blacktriangledown$  & \threeDIcon{}MN40 & \depthIcon{}SUN-D & \audioIcon{}ESC & \tactileIcon{}TAG-M & \eegIcon{}IN-EEG \\ \Xhline{3\arrayrulewidth}
I & 52.1    & 29.9     &  63.8  &  29.9   & 26.3      \\
T  & 48.3    &  47.6     &  59.4  &  71.9   & 39.0      \\
I+T &  65.4    & 50.9     &  71.2   & 63.6   &  35.9      
\end{tabular}
}
\caption{{\bf Align to different anchor data during training.} For different modalities, we show the classification results or zero-shot classification results when aligned to Image(I), Text(T) or Image and Text (I+T) during training.}\label{tab:supp_anchor_data}
\end{table}

\subsubsection{Different Ratio of Training Data}
\begin{figure}[h!]
\centering
\begin{tikzpicture}
    \begin{axis}[
        xtick={1, 2, 3, 4, 5, 6},
        xticklabels={1\%, 5\%, 25\%, 50\%, 75\%, 100\%}, 
        grid=both,
        grid style={line width=.1pt, draw=gray!10},
        major grid style={line width=.2pt,draw=gray!50},
        minor tick num=2,
        axis x line*=middle,
        axis y line*=left,
        height=1.9in,
        width=\linewidth,
        ylabel style= {align=center},
        ylabel={\threeDIcon{}O-LVIS Top-1},
        ylabel near ticks,
        yticklabel style = {font=\small},
        xticklabel style = {font=\small},
        legend style={at={(0.22,1.0)},anchor=north,cells={align=left}, font=\scriptsize},
        ybar,
        bar width=8pt,
        enlargelimits=0.15,
        ymin=0,
        ]

    \addplot[fill=gray!60] coordinates {
        (1, 3.4) %
        (2, 26.2) %
        (3, 36.1) %
        (4, 39.2)
        (5, 40.3)
        (6, 42.3)
    };
    \addlegendentry{OS-PointBERT~\cite{liu2023openshape}}
    
    \addplot[fill=pearDark] coordinates {
        (1, 20.7) %
        (2, 35.1) %
        (3, 48.2) %
        (4, 49.1)
        (5, 51.2)
        (6, 52.0)
    };
    \addlegendentry{\vitlensG}
    \end{axis}
\end{tikzpicture}
\caption{{\bf Using different ratios of training data} in OpenShape Triplets to train for 3D point cloud. Zero-shot prediction on O-LVIS is reported for OS-PointBERT and \vitlensG.}
\label{fig:supp_different_training_ratio}
\end{figure}
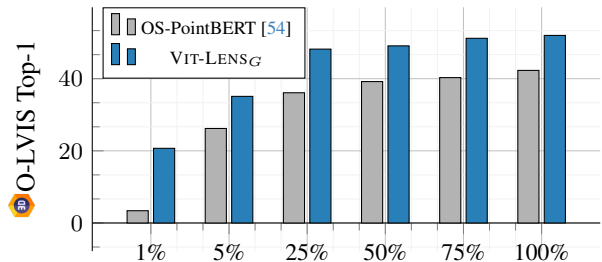
\noindent Our investigation delves into the influence of training data by performing ablation studies using different ratios of OpenShape Triplets~\cite{liu2023openshape} for training a 3D point cloud encoder. Specifically, we compare the performance of \vitlensG against OpenShape PointBERT in zero-shot classification on O-LVIS. The results, presented in ~\cref{fig:supp_different_training_ratio}, demonstrate that \vitlensG consistently outperforms OpenShape PointBERT across all ratios. 
Remarkably, in scenarios with limited training data (e.g., 1\% training data), \methodname showcases a significant performance advantage over PointBERT. This suggests the data-efficient nature of \methodname, attributed to the rich knowledge encapsulated within \ptvit.

\subsubsection{Additional Architectural Ablations}
This section focuses on additional ablation experiments centered around architectural designs. We focus on 3D tasks in this section. By default, our pretraining phase utilizes ULIP-ShapeNet Triplets~\cite{xue2023ulip}, followed by evaluation on the ModelNet40~\cite{wu2015modelnet} benchmark on zero-shot classification.

\par\head{Comparison with PointBERT.}
We conduct experiments to compare \methodname with PointBERT~\cite{yu2022pointbert}, a transformer based architecture for 3D point cloud understanding. This comparison involves aligning to the feature space of different CLIP variants and employing distinct pretraining datasets (ULIP-ShapeNet, ULIP2-Objaverse and OpenShape Triplets).
As is shown in~\cref{{tab:supp-abla-3d-pointBERT}}, \methodname outperforms PointBERT over all combinations of pretraining datasets and CLIP model for alignment.  This substantiates the efficacy of harnessing a pretrained \vit to advance 3D shape understanding.
\begin{table}[h]
\centering
\resizebox{0.8\linewidth}{!}{%
\begin{tabular}{l|c|c}
\multicolumn{1}{l|}{PT.D::CLIP Model $\blacktriangledown$} & PointBERT & \methodname \\ \Xhline{3\arrayrulewidth}
\dsA :: OpenAI-B16       &    60.2       &     61.7      \\
\dsA :: OpenCLIP-B16     &    62.6       &     65.4      \\
\dsA :: OpenAI-L14       &    61.2       &     63.3      \\
\dsA :: OpenCLIP-L14     &    65.4       &     70.6      \\
\dsB :: OpenAI-B16       &    70.6       &     73.4      \\
\dsB :: OpenCLIP-B16     &    71.7       &     74.8      \\
\dsB :: OpenAI-L14       &    74.1       &     76.1      \\
\dsB :: OpenCLIP-L14     &    77.8       &     80.6      \\
\dsC :: OpenCLIP-bigG14  &    84.4       &     87.4          
\end{tabular}
}
\caption{{\bf Comparisons with PointBERT.} We use different pretraining datasets(\dsA:ULIP-ShapeNet, \dsB:ULIP-2 Objaverse, \dsC:OpenShape Triplets) and different CLIP models as the foundation model for alignment. We report Top-1 accuracy on MN40.}\label{tab:supp-abla-3d-pointBERT}
\end{table}

\par\head{Configuration of Iter-CA-Attn type Lens in \methodname.} We delve into the impact of different design choices of the Iter-CA-Attn type employed in \methodname for 3D encoder. 
Our study encompasses the ablation of number of basis blocks (depth), as well as the exploration of parameter sharing beyond the second basis block (included), following~\cite{jaegle2021perceiver}. 
The results outlined in ~\cref{tab:supp-abla-perceiver-lens} indicate that, beyond a certain threshold, notably four in our setting, increasing the number of basis blocks does not yield improvements in performance. Moreover, parameter sharing among blocks demonstrates its capability to reduce parameters while achieving comparable performance. This emphasizes the efficacy and efficiency of the Iter-CA-Attn Lens architecture within \methodname for establishing connections between the 3D input and a \ptvit.
\begin{table}[h]
\centering
\resizebox{0.62\linewidth}{!}{%
\begin{tabular}{cc|c|c}
Depth & \begin{tabular}[c]{@{}c@{}}Share \\ Weights\end{tabular} & \#T.param & Acc@1 \\ \Xhline{3\arrayrulewidth}
2     &        -       & 34.1M    &  64.8     \\
4     &        \xmark  & 67.5M    &  64.2     \\
\rowcolor{lavenderweb}
4     &        \cmark  & 34.1M    &  65.4     \\
6     &        \xmark  & 100.8M   &  65.1     \\
6     &        \cmark  & 34.1M    &  64.0     \\
8     &        \xmark  & 134.2M   &  64.0     \\
8     &        \cmark  & 34.1M    &  64.3     \\
\end{tabular}
}
\caption{{\bf Configuration of Iter-CA-Attn Lens on depth and parameters sharing.} We show the number of trainable parameters and report the zero-shot Top-1 accuracy on MN40. The default setting is marked with color.
}\label{tab:supp-abla-perceiver-lens}
\vspace{-1em}
\end{table}

\par\head{Other hyper-parameters in \methodname.} 
We vary the number of latents used in the Lens of \vitlensB. Note that the number of latents equals to the sequence length of the \ptvit input. 
As delineated in~\cref{tab:supp-abla-hyperparam}, employing a larger number of latents, such as 384 and 512, shows slightly improved performance while concurrently increasing computational complexity measured in GFlops.  
This observation underscores the inherent capability of the CA-Iter-Attn type Lens to extract information from inputs of variable sizes and seamlessly connect them to the \ptvit, mitigating computational complexity.
Additionally, we investigate whether the inclusion of the \ptvit position embedding influences model performance. Specifically, we interpolate the original position embedding while varying the number of latents. The results presented in~\cref{tab:supp-abla-hyperparam} suggest that omitting the pretrained position embedding does not notably degrade performance. This suggests that the Lens is able to implicitly assimilate position information.
\begin{table}[h]
\centering
\resizebox{0.62\linewidth}{!}{%
\begin{tabular}{cc|c|c}
\#latents & ViT.pos & Flops & Acc@1 \\ \Xhline{3\arrayrulewidth}
   128       &   \xmark      &   54.0G   &  65.1   \\
   128       &   \cmark      &   54.0G   &  65.2   \\
   196       &   \xmark      &  75.4G    &  65.1   \\
   \rowcolor{lavenderweb}
   196       &   \cmark      &  75.4G   & 65.4    \\
   256       &   \xmark      &  94.6G   & 65.5    \\
   256       &   \cmark      &  94.6G   & 65.5    \\
   384       &   \xmark      &  136.4G  & 66.2    \\
   384       &    \cmark     &  136.4G  &  66.3   \\
   512       &    \xmark     &  179.5G  &  66.3   \\
   512       &    \cmark     &  179.5G  &  67.4  
\end{tabular}
}
\caption{{\bf Configuration of \#latents and ViT position embedding.} We vary the number of latent queries and switching the use of the original \ptvit position embeddings. The results showcase the corresponding GFlops to indicate computational complexity, along with reporting the Top-1 zero-shot accuracy on MN40. We show the  default setting marked with color for clarity.
}\label{tab:supp-abla-hyperparam}
\end{table}

\par\head{PointEmbed $\rightarrow$ Lens.} To validate the efficacy of the \ptvit, we investigate the performance of the ``PointEmbed $\rightarrow$ Lens'' paradigm. In this setup, the mean pooling feature of the CA-Iter-Attn Lens aligns directly with the CLIP feature space.
We conduct experiments with various hyper-parameter configurations, and the comprehensive outcomes are presented in~\cref{tab:supp-abla-point-perceiver}.
Specifically, the configuration featuring a ``depth of 6, with no parameter sharing'' possesses a total parameter count comparable to the default setting of \methodname (approximately 119M parameters). 
Despite having less trainable parameters, \methodname outperforms this variant of ``PointEmbed $\rightarrow$ Lens'' by a significant margin. Besides, \methodname also outperforms the rest variants.
This observation underscores the importance of harnessing the capabilities of the \ptvit.
\begin{table}[h]
\centering
\resizebox{0.9\linewidth}{!}{%
\begin{tabular}{ccc|cc|c}
Depth & \#latents & \begin{tabular}[c]{@{}c@{}}Share \\ Weights\end{tabular} & \#T.param & Flops & Acc@1 \\ \Xhline{3\arrayrulewidth}
 2 & 196       &   -           &  34.1M   & 27.4G  & 62.2  \\
 4 & 196       &   \xmark      &  67.5M  & 40.5G  &  62.4  \\
 8 & 196       &   \xmark      & 134.6M   & 66.7G  & 62.7  \\
 6 & 196       &   \xmark      & 101.2M   & 53.6G  & 61.9  \\
 6 & 196       &   \cmark      & 34.1M   & 53.6G  &  62.3 \\
 6 & 256       &   \xmark      & 101.3M   & 65.6G &  63.5  \\
 6 & 256       &   \cmark      & 34.2M   & 65.6G &   62.7 \\
 6 & 512  &    \xmark          & 101.5M    & 116.6G   &  62.5  \\
 6 & 512  &    \cmark          & 34.4M    &  116.6G &  62.3 \\ \midrule
 \multicolumn{6}{l}{\textit{Default setting} of \vitlensB} \\ \Xhline{2\arrayrulewidth}
 \rowcolor{lavenderweb}
 4 & 196  &    \cmark    & 34.1M  &  75.4G  &  65.4
\end{tabular}
}
\caption{{\bf Configurations for PointEmbed $\rightarrow$ Lens.} We vary the depth of Lens and alter sharing weights in Lens. We report the corresponding trainable parameters and zero-shot Top-1 accuracy on MN40. We show the  default setting marked with color at the bottom for clarity.
}\label{tab:supp-abla-point-perceiver}
\end{table}

\par\head{PointEmbed $\rightarrow$ \ptvit.} We also delve into the paradigm of ``PointEmbed $\rightarrow$ \ptvit''. As detailed in ~\cref{tab:supp-abla-point-ptvit}, training only the PointEmbed yields a zero-shot accuracy of 50\%, significantly lower than that achieved by \methodname due to the restricted number of trainable parameters.
Subsequently, enabling the training of transformer blocks results in an improved zero-shot performance. However, this specialized training approach tailored specifically for enhancing 3D understanding might limit the adaptability of the resulting \vit to other modalities, potentially impacting the overall generalization ability of the \vit.
In contrast, \methodname achieves commendable performance while largely preserving the core parameters of the \ptvit. This strategy effectively harnesses the extensive knowledge embedded within the \ptvit across diverse modalities, with only a marginal increase in new parameters, showcasing its robustness and adaptability.
\begin{table}[h]
\centering
\resizebox{1.\linewidth}{!}{%
\begin{tabular}{l|cc|c}
\begin{tabular}[c]{@{}c@{}} Unlocked Components in \vit \end{tabular} & \#T.param & Flops & Acc@1 \\ \Xhline{3\arrayrulewidth}
 None   &  7.3K  & 111.4G  &  50.0 \\
 \texttt{[CLS]}     &  7.3K  & 111.4G  &  53.6  \\
  \texttt{[CLS]}, \texttt{Proj}   & 1.1M   & 111.4G &  60.8  \\
  \texttt{[CLS]}, \texttt{Proj}, \texttt{Block.1}, \texttt{Block.2}    & 15.3M     & 111.4G   &  64.8     \\
  \texttt{[CLS]}, \texttt{Proj}, \texttt{Block.11}, \texttt{Block.12}  &  15.3M    & 111.4G   &  64.2    \\
  \texttt{[CLS]}, \texttt{Proj}, \texttt{Block.1} - \texttt{Block.4}    & 29.5M     & 111.4G   &  65.4    \\
  \texttt{[CLS]}, \texttt{Proj}, \texttt{Block.9} - \texttt{Block.12}    & 29.5M     & 111.4G   &  64.7    \\
  \texttt{[CLS]}, \texttt{Proj}, \texttt{Block.1} - \texttt{Block.6}    & 43.7M     & 111.4G   &   66.4   \\
  \texttt{[CLS]}, \texttt{Proj}, \texttt{Block.7} - \texttt{Block.12}    & 43.7M     & 111.4G   &  65.6    \\
  All & 86.6M  &  111.4G     &  67.7     \\ \midrule

  \multicolumn{4}{l}{\textit{Default setting} of \vitlensB} \\ \Xhline{2\arrayrulewidth}
  \rowcolor{lavenderweb}
  None(tune PointEmb, Lens) & 34.1M & 75.4G &  65.4
\end{tabular}
}
\vspace{-0.5em}
\caption{{\bf Configurations for PointEmbed$\rightarrow$\ptvit.} We vary the sub-modules of \ptvit unlocked during training. We report the corresponding trainable parameters, GFlops and zero-shot Top-1 accuracy on MN40. We show the  default setting marked with color at the bottom for clarity.
}\label{tab:supp-abla-point-ptvit}
\end{table}

\section{Further Discussion}
\label{sec:supp_discussion}
\begin{table}[bp]
\centering
\resizebox{.85\linewidth}{!}{%
\begin{tabular}{l|l|c}
Pretrained Data & Align to          & Acc@1 \\ \Xhline{3\arrayrulewidth}
ULIP-ShapeNet   & OpenCLIP-L14 (T)   & 48.7  \\
ULIP-ShapeNet   & Flan-T5 (T)        & 52.5  \\
ULIP-ShapeNet   & OpenCLIP-L14 (I+T) & 62.6  \\
ULIP2-Objaverse & OpenCLIP-L14 (T)   & 68.2  \\
ULIP2-Objaverse & Flan-T5 (T)        & 72.2  \\
ULIP2-Objaverse & OpenCLIP-L14 (I+T) & 79.0 
\end{tabular}
}
\vspace{-0.5em}
\caption{{\bf Train 3D encoder with pretrained Flan-T5 XL.} We use different pretrained data and foundation modelS for alignment. We report zero-shot Top-1 accuracy on MN40. 
}\label{tab:discussion-flant5}
\vspace{-0.5em}
\end{table}

\par\head{Beyond using \ptvit.}
The core of \methodname in advancing representations across diverse modalities relies on leveraging the profound knowledge embedded within the \ptvit. Given the significant enhancements facilitated by the \ptvit, an initial exploration involves employing the powerful Large Language Model (LLM) to encode inputs across various modalities. 
In this endeavor, we replace the \ptvit with Flan-T5 XL~\cite{chung2022flant5} within the \methodname architecture. To facilitate alignment, we introduce an additional trainable token. Training the model on ULIP-ShapeNet and ULIP2-Objaverse under various experimental configurations, we report the zero-shot classification performance on MN40.
Results are show in~\cref{tab:discussion-flant5}. Notably, when trained on ULIP-ShapeNet, the model exhibits proficient alignment with CLIP (I+T), achieving a notable top-1 zero-shot accuracy of 62.6\% on MN40.
Moreover, upon scaling the model to the ULIP2-Objaverse dataset enriched with textual captions, a remarkable improvement is observed. Specifically, it achieves an outstanding top-1 accuracy of 79\%, surpassing the performance obtained by training PointBERT from scratch with the same CLIP model for alignment. This outcome underscores the potential of this approach for omni-modal learning.
We leave further exploration of this promising avenue to future work.

{
    \small
    \bibliographystyle{ieeenat_fullname}
    \bibliography{main}
}


\end{document}